%% file: main.tex
\documentclass[preprint,12pt,authoryear]{elsarticle}

\usepackage{amssymb}
\usepackage{amsmath}

\usepackage{hyperref}
\usepackage{pbox}
\usepackage{subcaption}

\journal{Computers in Biology and Medicine}

\begin{document}

\begin{frontmatter}



\title{Structural-Based Uncertainty in Deep Learning Across Anatomical Scales: Analysis in White Matter Lesion Segmentation}

\author[chuv,hes,cibm]{Nataliia Molchanova\corref{cor1}}
\cortext[cor1]{Corresponding author.}
\ead{nataliia[dot]molchanova[at]unil[dot]ch}
\author[alta]{Vatsal Raina\fnref{label1}}
\fntext[label1]{Work done while at HES-SO Valais, Sierre, Switzerland}
\author[iso]{Andrey Malinin\fnref{label2}}
\fntext[label2]{Work done while at Yandex Research, Moscow, Russia}
\author[sinai]{Francesco La Rosa}
\author[chuv,hes]{Adrien Depeursinge}
\author[alta]{Mark Gales}
\author[unibas1,unibas2,unibas3]{Cristina Granziera}
\author[hes,unige]{Henning M\"uller}
\author[hes]{Mara Graziani}
\author[chuv,cibm]{Meritxell Bach Cuadra}

\affiliation[chuv]{
organization={Radiology Department, University of Lausanne and Lausanne University Hospital}, 
city={Lausanne},
country={Switzerland}
}
\affiliation[hes]{
organization={MedGIFT, Institute of Informatics, School of Management, HES-SO Valais-Wallis University of Applied Sciences and Arts Western Switzerland},
city={Sierre},
country={Switzerland}
}
\affiliation[cibm]{
organization={CIBM Center for Biomedical Imaging},
city={Lausanne},
country={Switzerland}
}
\affiliation[alta]{
organization={ALTA Institute, University of Cambridge}, 
city={Cambridge},
country={United Kingdom}
}
\affiliation[iso]{
organization={Isomorphic Labs},
city={London},
country={United Kingdom}
}
\affiliation[sinai]{
organization={Icahn School of Medicine at Mount Sinai}, 
city={New York City},
country={United States of America}
}
\affiliation[unibas1]{
organization={Translational Imaging in Neurology (ThINK) Basel, Department of Medicine and Biomedical Engineering, University Hospital Basel and University of Basel},
city={Basel},
country={Switzerland}
}
\affiliation[unibas2]{
organization={Department of Neurology, University Hospital Basel},
city={Basel},
country={Switzerland}
}
\affiliation[unibas3]{
organization={Research Center for Clinical Neuroimmunology and Neuroscience Basel (RC2NB), University Hospital Basel and University of Basel},
city={Basel},
country={Switzerland}
}
\affiliation[unige]{
{Department of Radiology and Medical Informatics, University of Geneva},
city={Geneva},
country={Switzerland}
}

\begin{abstract}
This paper explores uncertainty quantification (UQ) as an indicator of the trustworthiness of automated deep-learning (DL) tools in the context of white matter lesion (WML) segmentation from magnetic resonance imaging (MRI) scans of multiple sclerosis (MS) patients. Our study focuses on two principal aspects of uncertainty in structured output segmentation tasks. First, we postulate that a reliable uncertainty measure should indicate predictions likely to be incorrect with high uncertainty values. Second, we investigate the merit of quantifying uncertainty at different anatomical scales (voxel, lesion, or patient). We hypothesize that uncertainty at each scale is related to specific types of errors. Our study aims to confirm this relationship by conducting separate analyses for in-domain and out-of-domain settings. Our primary methodological contributions are (i) the development of novel measures for quantifying uncertainty at lesion and patient scales, derived from structural prediction discrepancies, and (ii) the extension of an error retention curve analysis framework to facilitate the evaluation of UQ performance at both lesion and patient scales. The results from a multi-centric MRI dataset of 444 patients demonstrate that our proposed measures more effectively capture model errors at the lesion and patient scales compared to measures that average voxel-scale uncertainty values. We provide the UQ protocols code at \url{https://github.com/Medical-Image-Analysis-Laboratory/MS\_WML\_uncs}.
\end{abstract}

\begin{graphicalabstract}
\includegraphics[width=\textwidth]{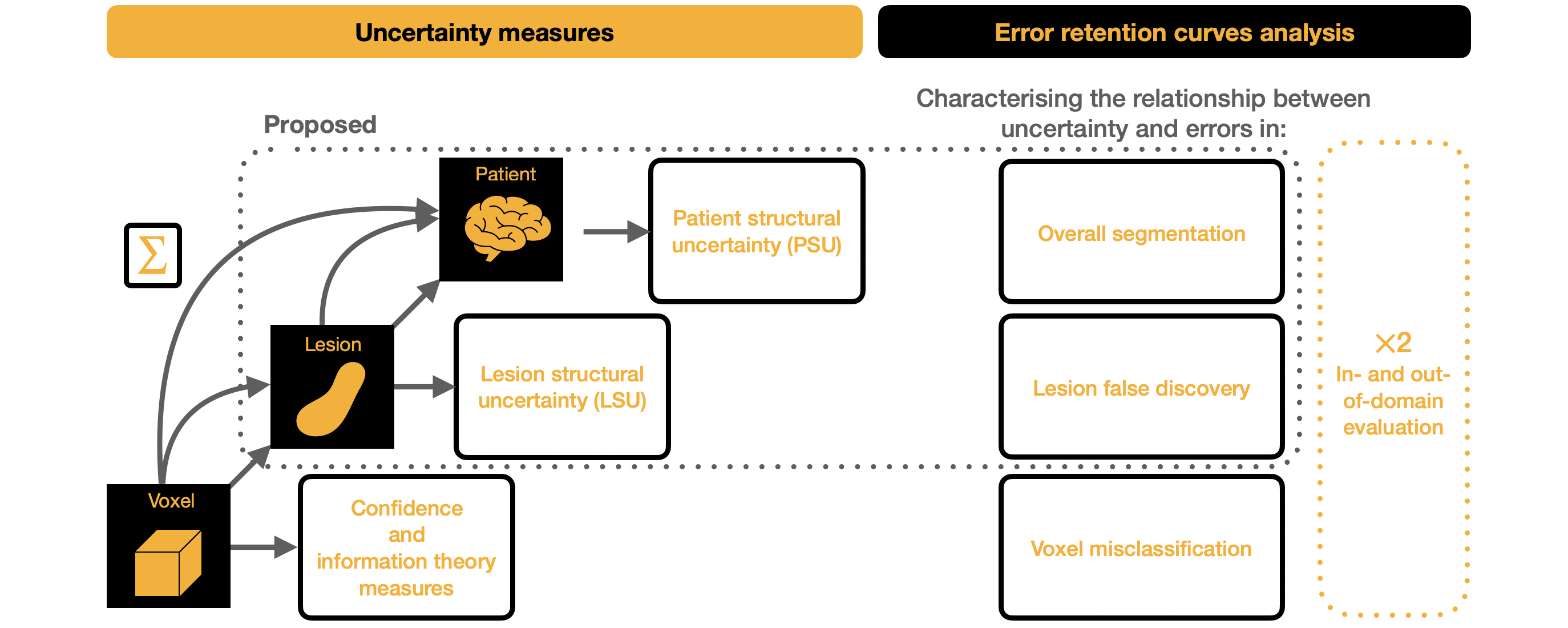}
\end{graphicalabstract}

\begin{highlights}
\item The proposed retention curve analysis enables a comparison of various uncertainty measures at three anatomical scales: voxel, lesion, and patient
\item The proposed lesion- and patient-scale uncertainty quantification measures suggest an advantage in identifying model errors in white matter lesion detection and segmentation
\item The proposed patient-scale uncertainty has a stronger correlation with Dice similarity score compared to state-of-the-art aggregation measures
\item The results are validated on a multi-center dataset with additional generalizability analysis on a cohort of patients with white matter hyperintensity 
\end{highlights}


\end{frontmatter}

\include{introduction}
\include{methods}

\include{results}
\include{conclusions}
\include{acknowledgements}

\appendix
\include{appendix}

\bibliographystyle{elsarticle-harv} 
\bibliography{cas-refs}

\end{document}

%% file: introduction.tex
\section{Introduction}

Multiple sclerosis (MS) is a chronic, progressive autoimmune disorder of the central nervous system affecting approximately 2.8 million people worldwide \citep{prevalence}. The primary characteristics of MS are demyelination, axonal damage, and inflammation due to the breakdown of the blood-brain barrier \cite{reich_multiple_2018, mac}. The diagnostic criteria for MS include both neurological symptoms observation and magnetic resonance imaging (MRI) examination for the presence of lesions disseminated in time and space \citep{mac, hemond_magnetic_2018, wattjes}. White matter lesions (WMLs) are a hallmark of MS, indicating the regions of inflammation in the brain, typically assessed through FLAIR or T1-weighted modalities \citep{flair, hemond_magnetic_2018}. On FLAIR scans, WMLs are visible as hyperintense regions with periventricular area, brainstem, and spinal cord being prevalent lesion sites. The size, shape, and count of WMLs vary markedly across patients. While crucial for diagnosis and monitoring, the manual annotation of new and enlarged lesions is a time-consuming and skill-demanding process.

The task of automated WML segmentation has propelled the development of novel image processing techniques for many years \citep{sotamsseg, LLADO2012164}. More recently, algorithms have been boosted by the success of deep learning (DL) in computer vision. DL methods quickly became state-of-the-art for WML segmentation, providing better performance at faster processing times at faster processing times \citep{mssegreview, SPAGNOLO2023103491}. Various DL models were explored in application to WML segmentation, with U-Net architecture being the most common model at faster processing times \citep{SPAGNOLO2023103491}.

The potential clinical application of DL methods raises safety concerns. These include the black-box nature of such approaches and their susceptibility to variations in test data, known as domain shifts \citep{doi:10.1148/ryai.2020190043}. Additionally, common factors such as limited data availability, imperfect annotations, and ground-truth ambiguity due to inter-rater variability further challenge the reliability of DL model predictions, potentially hindering their seamless integration into clinical practice \citep{osti_1561669}. The field of uncertainty quantification (UQ) offers a possibility to tackle this issue by estimating the ``degree of untrustworthiness'' of model predictions \citep{osti_1561669}, focusing on two main uncertainty sources \citep{uncs_survey}: i) data noise, captured by data uncertainty, and ii) training data scarcity or domain shifts, captured by model uncertainty. In the context of high-risk AI applications, the information about the trustworthiness of model predictions is important not only from an engineering perspective, but also for the end-users, \textit{e.g.} clinicians \citep{graziani-2022}.

Consequently, UQ is gaining popularity within the field of medical image analysis not only as a way to assess prediction trustworthiness. However, the usage of uncertainty extends beyond quality control to accommodate such applications as improving prediction quality, domain adaptation, active learning, and other applications \citep{uncs_survey, doi:10.1148/radiol.222217, ZOU2023100003, lambertreview}. In medical image segmentation tasks, uncertainty is usually assessed by treating semantic segmentation as pixel or voxel classification, computing uncertainty for each pixel or voxel prediction. Given the structure of a segmentation model output, it is also possible to explore uncertainty values associated with some region of prediction. Several works explore uncertainty associated with a segmented region of interest, \textit{e.g.} structure- or lesion-wise \citep{ROY201911,10.3389/fncom.2019.00056,rottmann2019prediction, TAlber, Dojat}, or for a whole prediction on a patient \citep{10.3389/fnins.2020.00282, Whitbread_2022}.

\subsection{Related works on uncertainty quantification in multiple sclerosis}

Prior research on UQ for WML segmentation explored different techniques, including single-network deterministic methods \citep{MCKINLEY2020102104, Dojat}, Monte Carlo Dropout (MCDP) \citep{TAlber}, batch ensembles \citep{Dojat}. Our previous study \citep{shifts20} investigated the deep ensembles \citep{NIPS2017_9ef2ed4b} and compared them with the MCDP method \citep{mcdp}, showing the advantage of the first one. The utility of a specific UQ method depends on a particular application and available resources \citep{lambertreview, uncs_survey, doi:10.1148/radiol.222217, ZOU2023100003}. Deep ensembles were subsequently shown to have a higher quality of uncertainty estimates compared to other methods, while being computationally less effective compared to single-shot models or batch ensembles \citep{lambertreview, uncs_survey, doi:10.1148/radiol.222217, ZOU2023100003}. The deep ensemble is a deterministic method as the inference of each member is; thus, the reliability of this UQ method can be studied without a concern about the repeatability of the results.

Using ensemble methods or sampling UQ methods, based on obtaining samples from the posterior distribution, allows for the exploration of various uncertainty measures.  Several measures of voxel-scale uncertainty have been explored, including variance, entropy, mutual information \citep{TAlber, DojatMICCAI}. Our previous study expanded this list by exploring a common negated confidence and more advanced measures of model uncertainty, such as reverse mutual information and expected pairwise Kullback–Leibler divergence \citep{shifts20, molchanova_isbi}. Several studies with different UQ methods and measures used, observe that voxel scale uncertainty tends to be the highest at the borders of WMLs, especially larger ones \citep{MCKINLEY2020102104, TAlber, shifts20, DojatMICCAI, molchanova_isbi}, resembling partial-volume \citep{FARTARIA2018245, FARTARIA2019101938} or inter-rater disagreement maps. 

In MS, some works explored uncertainty associated with a segmented region of interest, \textit{i.e.} at the lesion scale \citep{TAlber, DojatMICCAI, molchanova_isbi}. The pioneering study \citep{TAlber} suggested computing a log-sum of voxel-scale uncertainties across a predicted lesion region, using different voxel-scale uncertainty maps.  Analogously, mean average voxel uncertainty values across the lesion region were explored \citep{DojatMICCAI}. \citet{DojatMICCAI} showed the advantages of structural UQ based on graph neural networks over voxel aggregation methods. Our prior research \citep{DojatMICCAI} demonstrated that lesion-scale uncertainty, computed through disagreement in structural predictions, is more effective at identifying false-positive lesions than aggregating voxel-scale uncertainties. Although we explored advanced measures such as expected KL divergence and reverse mutual information \citep{malinin2020Struct}, they did not exhibit any significant advantage over the more commonly employed entropy and mutual information in medical image analysis. In the context of MS lesion segmentation, the patient-scale uncertainty remains less explored.

Besides these various measures, prior works proposed different ways to compare uncertainty measures. Ideally, a high uncertainty score should highlight the predictions that are most likely to be wrong. Hence, we expect a reliable uncertainty measure to reflect the increased likelihood of an erroneous prediction and thus correlate with model mistakes. For classification tasks, a calibration of uncertainty is measured to assess its quality, similarly the uncertainty quality can be compared at the voxel scale. At the lesion-/ patient- scales the calibration metrics are not explicitly defined. When investigating lesion-scale measures, \citet{TAlber} looked into uncertainty-based prediction filtering as a means to correlate uncertainty and false positive errors, and \citet{DojatMICCAI} used accuracy-confidence curves. Our previous work redefines an error retention curve analysis to quantify the relationship between uncertainty and lesion detection errors \citep{molchanova_isbi}. Prior to that the error retention curve analysis has been explored to compare classification or segmentation pixel-/voxel-scale uncertainty measures for various tasks as a way to quantify its relationship with an error / quality metric of a choice \citep{malinin2019uncertainty, malinin2020Struct, brats}. This is a necessary analysis for various practical clinical implementations, including a signaling uncertainty-based system to warn medical specialists about the potential errors in automatic predictions, automatic uncertainty-based filtering of errors, or active learning where the hardest, \textit{i.e.} most likely erroneous examples need to be selected.

Various studies on UQ for WML segmentation use similar U-net-like deep learning models \citep{ronneberger-2015, cicek-2016, TAlber, shifts20, DojatMICCAI}, which have been widely explored in application to the MS lesion segmentation task \citep{mssegchallenge, sotamsseg, LAROSA2020102335, SPAGNOLO2023103491}. While there is an agreement about the DL model, studies were conducted on various datasets, predominantly private ones. There had not been a public benchmark dataset for the UQ methods evaluation within the context of WML segmentation before the Shifts 2.0 Challenge \citep{shifts20}.

\subsection{Our contributions}

This study extends our previous work \citep{molchanova_isbi} and introduces advancements in uncertainty quantification (UQ) methods, focusing on MRI segmentation across voxel, lesion, and patient scales. We introduce a novel patient-scale uncertainty measure that leverages ensemble member disagreement to more accurately identify segmentation errors. To compare patient-scale measures, we redefine the error retention curve analysis, enabling a better understanding of their performance in detecting poor segmentation quality. Our quantitative evaluation is conducted in both in-domain and out-of-domain settings using a total of 404 scans to mirror the diversity of MRI data coming from several studies, medical centers, and scanners. Additionally, this research provides a comparison of uncertainty measures across different anatomical scales, highlighting their capacity to detect voxel misclassification, lesion false discovery, and general segmentation inaccuracies, considering clinically relevant applications. The proposed UQ framework is specifically tailored for WML segmentation on FLAIR MRI scans. Through additional evaluation, we confirm the generalizability of a similar task of white matter hyperintensity segmentation on 2D FLAIR MRI scans.

Our contributions include:
\begin{itemize}
    \item Proposing the error retention curves analysis for instance-detection tasks, enabling an evaluation of lesion-scale UQ methods in their ability to capture lesion false detection errors.
     \item Proposing a patient-scale uncertainty measure, a novel approach for WML segmentation evaluation, enhancing the understanding of overall segmentation failure.
     \item Proposing the extension of the error retention curves analysis for patient-scale to compare the ability of different uncertainty measures to capture overall segmentation quality.
\end{itemize}

%% file: methods.tex
\section{Materials and methods}

\subsection{Data}

\begin{figure}
    \centering
    \includegraphics[width=\linewidth]{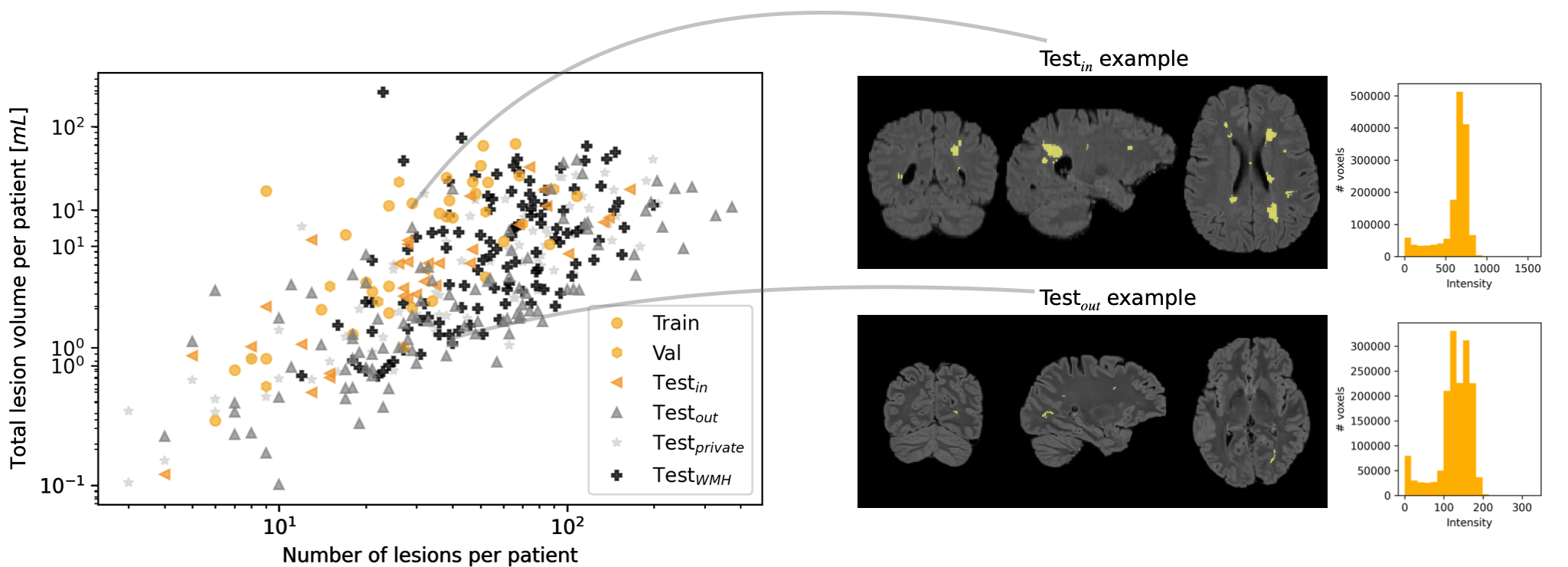}
    \caption{Illustration of the domain shift between the in-domain datasets (Train, Val, Test$_{in}$) and the out-of-domain dataset (Test$_{out}$, Test$_{private}$, and Test$_{WMH}$) brought by the differences in the MS stages and medical centers. On the left, the plot of the total lesion volume in milliliters versus the number of lesions per scan for in-domain (orange) and out-of-domain (gray and black) sets reveals the difference in the lesion load (as a proxy to an MS stage) between different domains. On the right, typical examples from the Test$_{in}$ and Test$_{out}$ sets illustrate the difference in the lesion load, as well as the intensity differences brought by the change of the medical center (\textit{i.e.} scanner, technicians, annotators, and other parameters contributing to the domain shift) and MS stages (\textit{i.e.} smaller lesion load and size).}
    \label{fig:data}
\end{figure}

The initial study creating the data was designed as a part of the Shifts 2.0 Challenge \citep{shifts20} specifically for the exploration of uncertainty quantification across shifted domains. This configuration comprises three publicly available datasets and a single private one. Data is separated into in-domain (Train, Val, Test$_{in}$) and out-of-domain (Test$_{out}$) subsets. This enables UQ evaluation both with and without the domain shift. Data split into in- and out-of-domain sets is designed to maximize the drop of model performance in lesion segmentation in the out-of-domain test. From a clinical perspective, the domain shift is provided by the difference in medical center, scanner, annotators, and MS stages (Table \ref{tab:data}). The Test$_{in}$ and Test$_{out}$ show a prominent difference in lesion distributions likely brought by the differences of MS stages distributions (see Figure \ref{fig:data}).

We extend this existing public benchmark by including a large in-house dataset (Test$_{private}$, 162 patients) collected in the Basel University Hospital, Switzerland \citep{granziera2018}. While Test$_{private}$ should be treated as an out-of-domain, the lesion profiles overlap with both Test$_{in}$ and Test$_{out}$ (see Figure \ref{fig:data}).

For the additional assessment of generalizability and repeatability, we add an evaluation on a similar task of white matter hyperintensity (WMH) segmentation. We use a publicly available test set from the WMH Segmentation Challenge \citep{kuijf2022} comprising 110 subjects. On MRI FLAIR scans, WMH has a similar WML MS visual representation, but not localization \citep{flair}. WMHs come from a different pathology related to vascular abnormalities rather than MS \citep{erten-lyons-2013}. The WMH Segmentation Challenge dataset contains 2D FLAIR scans with 3mm thickness, compared to 0.8-2.2 mm slice thickness in the rest of the datasets. The lack of information in the \(z\)-axis contributes to the domain shift in addition to differences in study, medical center, underlying pathology, annotation protocol, among others. Additionally, this cohort exhibits higher lesion loads and larger lesion sizes (see Figure \ref{fig:data}). 

For WML and WMH segmentation, this study uses FLAIR MRI scans and their manual WML annotations. FLAIR scans from Test$_{private}$ and Test$_{WMH}$ underwent a common pre-processing pipeline similar to the Shifts 2.0 Challenge pre-processing, including skull stripping \citep{isensee2019}, bias field correction \citep{tustison2010}, and interpolation to 1mm isovoxel space. 
Information about data sources, metadata, and data splits is provided in Table \ref{tab:data}. Figure \ref{fig:data} illustrates some differences between domains brought by variations in MS stage distributions and scanner changes, affecting the lesion characterization and intensity features, respectively. Other factors, such as changes in study design, lesion annotators, scanner operators, may also contribute to the domain shift.

\begin{table}[htbp]
    \centering
    \tiny
    \begin{tabular}{|*{6}{c|}c|}
    \hline
    \textbf{Domain} 
        & \multicolumn{3}{c|}{\textbf{In-domain}} 
        & \multicolumn{2}{c|}{ \pbox{3cm}{\centering \textbf{Out-of-domain MS}}} 
        &  \pbox{2cm}{\centering \textbf{Out-of-domain WMH}} \\\hline
    
    \textbf{Source} 
        & \multicolumn{3}{c|}{\pbox{4cm}{\centering \citet{CARASS201777}, \newline \citet{commowick2018}}} 
        & \pbox{1.5cm}{\centering \citet{Lesjak2017ANP}, \newline \citet{advanced}} 
        & \pbox{1.5cm}{\centering \citet{granziera2018} }
        &\pbox{2cm}{\centering \citet{kuijf2022} } \\[2ex]
    
    \pbox{2cm}{\centering \textbf{Medical center location}}
        & \multicolumn{3}{c|}{\pbox{4cm}{\centering Rennes, Bordeaux and Lyon (France), Best (Netherlands)}}
        & \pbox{1.5cm}{\centering Ljubljana (Slovenia), Lausanne (Switzerland)} 
        &  \pbox{1.5cm}{\centering Basel (Switzerland)}
        & \pbox{2cm}{\centering Utrecht and Amsterdam (Netherlands), Singapore} \\[2ex]
    
    \textbf{Scanners} 
        & \multicolumn{3}{c|}{\pbox{3cm}{\centering Siemens (Aera 1.5T, Verio 3.0T), GE Disc 3.0T, Philips (Ingenia 3.0T, Medical 3.0T)}} 
        & \pbox{1.5cm}{\centering Siemens Magnetom Trio 3.0T}
        & \pbox{1.5cm}{\centering Siemens Magnetom Prisma 3.0T}
        & \pbox{2cm}{\centering 3T Philips Achieva, Siemens TrioTim 3.0T, Philips Achieva 3.0T, Ingenuity 3.0T, GE Signa (1.5T, 3.0T)}
    \\[2ex]
    
    \textbf{M:F ratio range} 
        & \multicolumn{3}{c|}{0.21-0.4}
        & 0.23-0.70
        & 0.68 
        & - \\
        
    \textbf{MS stages}
    & \multicolumn{3}{c|}{RR, PP, SP}
    &  \pbox{1.5cm}{\centering CIS, RR, SP, PP}
    & RR, PP, SP 
    & - \\
    
    \textbf{\# raters} 
    & \multicolumn{3}{c|}{2 / 7}
    & consensus / 3 
    & consensus  
    & consensus \\
    
    \pbox{2cm}{\centering \textbf{Inter-rater agreement (Dice score)}}
    & \multicolumn{3}{c|}{0.63 and 0.71} 
    & 0.78 and - 
    & - 
    & - \\\hline
    
    \textbf{Set name} 
    & Train 
    & Val 
    & Test$_{in}$ 
    & Test$_{out}$ 
    & Test$_{private}$ & 
    Test$_{WMH}$ \\\hline
    
    \textbf{\# scans}  & 33  & 7  & 33   & 99 & 162 & 110 \\
    \pbox{2cm}{\centering \textbf{$\#$ lesions per scan, Q2 (IQR)}} & 34 (20-50)  & 26 (19-61)    & 30 (15-47)   & 39 (20-77)   & 63 (25-88)   & 60 (37-83) \\
    
    \pbox{2cm}{\centering \textbf{Total lesion volume per scan, Q2 (IQR) [mL]}} & 12.5 (3.1-27.8) & 15.5 (4.0-24.7) & 7.2 (3.7-11.3) & 2.7 (1.3-7.3) & 7.4 (2.4-14.3) & 9.4 (3.3-20.3) \\\hline
    
    \end{tabular}
    
    \caption{
    Data splits and meta information. MS stages are clinically isolated syndrome (CIS), relapsing remitting (RR), primary progressive (PP), and secondary progressive (SP). Computed statistics are median (Q2) and interquartile range (IQR). Computed statistics are median (Q2) and interquartile range (IQR).
    }
    \label{tab:data}
\end{table}

\subsection{Uncertainty quantification}

This work implements deep ensembles \citep{NIPS2017_9ef2ed4b} for UQ by training multiple networks with identical architecture but different random seed initializations. The random seed controls several factors, for instance, weights initialization, training sample selection, random augmentations, and stochastic optimization algorithms. Although each ensemble member has distinct model weights, they all stem from the same posterior distribution. This causes varied predictions among ensemble members for the same input example. The spread or variation in these predictions serves as an uncertainty estimate.

\subsubsection{Uncertainty quantification at different anatomical scales}

In an image segmentation task, a class prediction is not a single value but an image-size map. Thus, the disagreement between the ensemble members can be quantified not only for each voxel of the prediction but also for a subset of its elements. For WML segmentation, the model prediction is a 3D probability map. We can quantify the uncertainty associated with the decision taken in each voxel, thus obtaining another 3D map with voxel-scale uncertainty values. We can also quantify uncertainty associated with a set of predictions within a region of a particular lesion, thus obtaining an uncertainty score for each predicted lesion. Similarly, we can quantify uncertainty for the whole patient. We implement several uncertainty measures at each anatomical scale (voxel, lesion, or patient). The exact mathematical formulation for the previous existing and proposed UQ measures are summarized in Table 2 and described hereafter.

\paragraph{Voxel-scale uncertainty measures} Perceiving segmentation as a classification of each voxel of an image, one could use uncertainty measures available for classification tasks to quantify uncertainty for per-voxel predictions. The common uncertainty measures in this case will be negated confidence and information theory measures such as entropy of expected, expected entropy, or mutual information which respectively depict different \emph{total}, \emph{data}, and \emph{model} uncertainty.

\paragraph{Lesion-scale uncertainty measures} Given a WML segmentation task, we can compute a single uncertainty score for each predicted connected component, \textit{i.e.} lesion. Differently from previous measures that aggregate voxel-scale uncertainties \citep{TAlber, Dojat}. Our previous work \citep{molchanova_isbi} proposes a novel lesion-scale uncertainty defined directly through the disagreement between the lesion structural predictions of ensemble members. We hypothesize that looking at the disagreement in structural predictions, \textit{i.e.} predicted lesion regions, might be more beneficial for the discovery of false positive lesions.

To define our proposed measure, we consider the ensemble of \( M \) models, where each member model is parametrized by weights \( \theta_m \), \( m \in \{ 0, 1, \dots, M-1 \} \). The ensemble probability prediction is obtained by computing a mean average across members. Then, the binary lesion segmentation mask is obtained by applying a threshold \( \alpha \) to the softmax ensemble prediction, where \( \alpha \) is chosen based on the Dice similarity coefficient maximized on the validation dataset. Analogously, by applying the threshold \( \alpha \) to the softmax predictions of each of the ensemble models, we can obtain the binary lesion segmentation masks predicted by each model \( m \) in the ensemble. Let \( L \) be a \emph{predicted lesion} that is a connected component from the binary segmentation map obtained from the ensemble model; and \( L^m \) is the \emph{corresponding lesion} predicted by the model \( m \), determined as the connected component on the binary segmentation map predicted by the \( m \)-th member with maximum intersection over union (IoU) with \( L \). If the softmax probability threshold is optimized for each member model separately based on the highest Dice score, the resulting thresholds will be different from \( \alpha \) and will be member-specific: \( \alpha^m \), \( m \in \{ 0, 1, \dots, M-1 \} \), instead of \( \alpha \). Then, the binary segmentation maps obtained with \( \alpha^m \) will lead to different corresponding lesion regions, called \( L^{m,+} \). Then, the proposed measure, lesion structural uncertainty (LSU), is defined as follows:

\begin{equation}
    LSU=1 - \frac{1}{M} \sum\limits_{m=0}^{M-1} IoU(L, L^m),
    \label{eq:lsu}
\end{equation}

and 

\begin{equation}
    LSU^+=1 - \frac{1}{M} \sum\limits_{m=0}^{M-1} IoU(L, L^{m,+}).
    \label{eq:lsu_plus}
\end{equation}

\paragraph{Patient-scale uncertainty measures} Patient-scale uncertainty offers the most compact way of uncertainty representation considering the clinical practice, that is presenting a single uncertainty score per patient. Analogously to the lesion scale, the patient-scale uncertainty can be computed by averaging voxel or lesion uncertainties.
Using similar reasoning as for the lesion scale, we propose a patient-scale measure analogous to LSU (Equation \ref{eq:lsu}), where instead of the lesion region \( L \) the total segmented lesion region is used.
To define these measures, let S be a set of voxels predicted as lesion class by the ensemble model, \( S^m \) - set of voxels predicted as lesion class by the \( m \)-th member model in the ensemble, and \( S^{m,+} \) is the same, but obtained with the member-specific threshold \( \alpha^m \). Then, the proposed patient structural uncertainty measures are defined as:

\begin{equation}
    PSU=1 - \frac{1}{M} \sum\limits_{m=0}^{M-1} IoU(S, S^m),
    \label{eq:psu}
\end{equation}

and 

\begin{equation}
    PSU^+=1 - \frac{1}{M} \sum\limits_{m=0}^{M-1} IoU(S, S^{m,+}).
    \label{eq:psu_plus}
\end{equation}

\begin{table}[!h]
\centering
\tiny
    \begin{subtable}{\textwidth}
        \centering
        \caption{\textbf{Voxel-scale uncertainty measures} computed for each pixel $i \in B$ of an input scan $\mathbf{x}$ ($B$ is a set of voxels defining the brain region), $\mathbf{y}$ - targets, $c \in \{ 0, 1, \dots, C-1 \}$ is the class ($C=2$ for binary segmentation), $P(y_i=c|\mathbf{x}, \boldsymbol \theta^{m})$ is a softmax probability predicted by the $m$-th member in the ensemble of $M$ models, and $\hat P(y_i=c|\mathbf{x}) = \frac{1}{M} \sum\limits_{m=0}^{M-1} P(y_i=c|\mathbf{x}, \boldsymbol \theta^{m})$ is the probability predicted by ensemble.}
        \begin{tabular}{|c|}
            \hline
            Negated confidence \\
            $NC_i=-\operatorname*{argmax}\limits_{c=0,..,C-1} \frac{1}{M}\sum\limits_{m=0}^{M-1} P(y_i=c|\mathbf{x}, \boldsymbol \theta^{m})$ \\[2ex]
            \hline
            Entropy of expected \\
            $EoE_i=-\sum\limits_{c=0}^{C-1} \hat P(y_i=c|\mathbf{x})log\hat P(y_i=c|\mathbf{x})$ \\[2ex]
            \hline
            Expected entropy \\
            $ExE_i=-\frac{1}{M}\sum\limits_{m=0}^{M-1} \sum\limits_{c=0}^{C-1} P(y_i=c|\mathbf{x}, \boldsymbol \theta^{m})\log P(y_i=c|\mathbf{x}, \boldsymbol \theta^{m})$ \\[2ex]
            \hline
            Mutual information \\
            $MI_i=EoE_i-ExE_i$ \\
            \hline
        \end{tabular}
    \end{subtable}%
    \hfill
    \begin{subtable}{\textwidth}
        \centering
        \caption{\textbf{Lesion-scale uncertainty measures} computed for each \textit{predicted lesion} $L$, that is a connected component on the predicted binary segmentation map. The last is obtained by applying a threshold $\alpha$ to the softmax ensemble prediction $\hat P(\mathbf{y}=\mathbf{1}|\mathbf{x}) = \frac{1}{M} \sum\limits_{m=0}^{M-1} P(\mathbf{y}=\mathbf{1}|\mathbf{x}, \boldsymbol \theta^{m})$, where $\alpha$ is chosen based on the Dice similarity coefficient maximized on the validation dataset. $L^m$ is the corresponding lesion predicted by the $m^{th}$ member model, determined as the connected component on the binary segmentation map predicted by the $m^{th}$ member (threshold $\alpha$ applied to $P(\mathbf{y}=\mathbf{1}|\mathbf{x}, \boldsymbol \theta^{m}),  m \in \{ 0, 1, \dots, M-1 \}$) with maximum intersection over union (IoU) with $L$. If the softmax probability threshold is optimized based on the highest Dice score for each member model separately, the resulting thresholds will be different from $\alpha$ and will be member-specific: $\alpha^{m},  m \in \{ 0, 1, \dots, M-1 \} $ instead of $\alpha$. Then, the binary segmentation maps obtained by applying $\alpha^{m}$ to $P(\mathbf{y}=\mathbf{1}|\mathbf{x}, \boldsymbol \theta^{m}),  m \in \{ 0, 1, \dots, M-1 \}$ will lead to different corresponding lesion regions, called $L^{m,+}$.}
        \begin{tabular}{|c|}
            \hline
            Voxel uncertainties aggregation via mean average  \\
            $\overline{EoE}_{L}=\frac{1}{|L|}\sum\limits_{{i} \in L} EoE_i,$ \\[2ex]
            Analogously, $\overline{ExE}_{L}, \overline{NC}_{L}, \overline{MI}_{L}$  are defined. \\[1ex]
            \hline
            \textbf{Proposed} lesion structural uncertainty (LSU) \\
            $LSU=1 - \frac{1}{M} \sum\limits_{m=0}^{M-1} IoU(L, L^m)$ \\
            and \\
            $LSU^+=1 - \frac{1}{M} \sum\limits_{m=0}^{M-1} IoU(L, L^{m,+})$ \\[2ex]
            \hline
        \end{tabular}
    \end{subtable}%
    \hfill
    \begin{subtable}{\textwidth}
        \centering
        \caption{\textbf{Patient-scale uncertainty measures} computed for patient. $S$ is a set of voxels predicted as lesions by the ensemble model, $S^{m}$ is a set of voxels predicted as lesions by the model $m$, and $S^{m,+}$ is the same, but obtained with the member-specific threshold $\alpha^{m}, m=0,1,..., M-1$. $W$ - set of lesions predicted by the ensemble model.}
        \begin{tabular}{|c|}
            \hline
            Voxel uncertainties aggregation via mean average  \\
            $\overline{EoE}_{B}=\frac{1}{|B|}\sum\limits_{{i} \in B} EoE_i,$ \\[2ex]
            Analogously, $\overline{ExE}_{B}, \overline{NC}_{B}, \overline{MI}_{B}$  are defined. \\[1ex]
            \hline
            {Proposed} lesion uncertainties aggregation via mean average  \\
            $\overline{LSU}=\frac{1}{|W|}\sum\limits_{{l} \in W} LSU_l,$ \\[2ex]
            Analogously, $\overline{LSU^+}$  is defined. \\[1ex]
            \hline
            \textbf{Proposed} patient structural uncertainty (PSU) \\
            $PSU=1 - \frac{1}{M} \sum\limits_{m=0}^{M-1} IoU(S, S^m)$ \\
            and \\
            $PSU^+=1 - \frac{1}{M} \sum\limits_{m=0}^{M-1} IoU(S, S^{m,+})$ \\[2ex]
            \hline
        \end{tabular}
    \end{subtable}
    \caption{Definitions of uncertainty measures at three anatomical scales: voxel, lesion, and patient.}
    \label{table:uncs_def}
\end{table}

\subsection{Quantitative evaluation of uncertainty measures}
\label{sec:evaluation}

Uncertainty has a relation to errors made by a model: ideally, a higher uncertainty expresses an increased likelihood of erroneous prediction. For each of the anatomical scales: voxel, lesion, and patient, the ``error'' definition can vary. For example, a voxel-scale error can be simply defined as a voxel misclassification, a lesion-scale error can be defined as a lesion misdetection, and a patient-scale error can be a summary of voxel errors. In this work, we want to compare voxel-, lesion-, and patient-scale uncertainty measures in terms of their ability to capture errors of different kinds. For this, we use an error retention curve analysis \citep{malinin2019uncertainty, shifts20, brats}, previously introduced only for voxel-scale uncertainty, and extended for lesion and patient scales in this work.

\subsubsection{Error and quality metrics}
\label{sec:error}

We start by defining errors on the voxel and lesion scale as well as quality metrics used in this work for model performance characterization and error retention curve analysis.

\paragraph{Voxel-scale errors} Similarly to a classification task, the errors at the voxel scale will include false positives and negatives (FP and FN, respectively). Based on FP, FN, true positives (TP), and true negatives (TN), one derives metrics like true positive rate (TPR) and positive predictive value (PPV), which measure correctly classified voxels against ground truth or predicted lesions, respectively. To evaluate both error types, we use the F$_1$ score, also known as the Dice similarity score (DSC) in image processing. However, it is well known that the DSC metric suffers from a bias to the occurrence rate of the positive class, \textit{i.e.} lesion load, jeopardizing the comparison of results. We thus additionally utilize the normalized DSC (nDSC) \citep{Vatsal_isbi} for the model evaluation. In a nutshell, nDSC scales the precision at a fixed recall rate to tackle the lesion load bias.

\paragraph{Lesion-scale errors} Analogously, true positive, false positive, and false negative lesions (TPL, FPL, FNL) can be defined if the criteria for lesion (mis)detection are given. While some studies accept minimal overlap for detection \citep{TAlber, CARASS201777, LAROSA2020102335}, we apply a 25\% intersection over the union threshold for a predicted lesion to be considered a TPL. For the FNL definition, we consider a zero overlap with the prediction. A FNL is a ground truth lesion that has no overlap with predictions. Metrics derived from TPL, FPL, and FNL include Lesion TPR, PPV, and F$_1$, further referred to as LTPR, LPPV, LF$_1$. The differences at the voxel scale include: i) uncertainty cannot be quantified for FNLs, as they are not predicted lesions; ii) it is not possible to define a true negative lesion. The metrics definitions can be found in \ref{appendix:metrics}.

\subsubsection{Error retention curve analysis}

\begin{figure}[ht]
    \centering
    \includegraphics[width=0.55\textwidth]{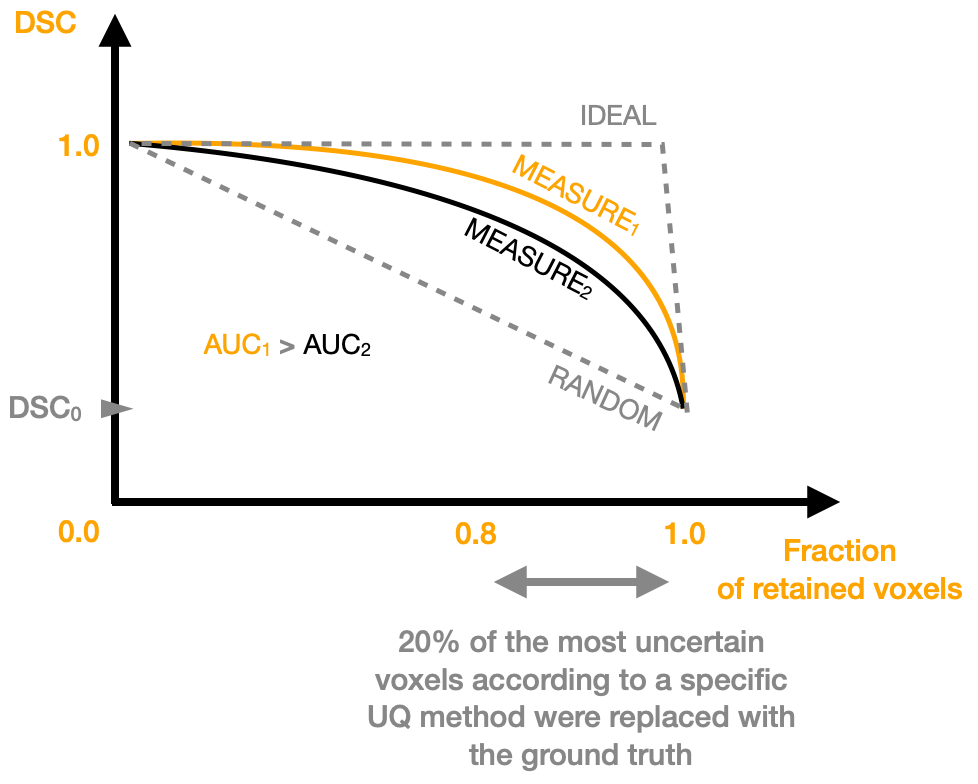}
    \caption{An illustration of a Dice score retention curve (DSC-RC) for assessing the correspondence between voxel uncertainty (MEASURE$_1$ and MEASURE$_2$) and segmentation quality measured by DSC. DSC$_0$ - quality of the predicted segmentation before voxel replacement. IDEAL and RANDOM RCs are built for the ideal and random uncertainty and are the upper and lower bounds of the uncertainty-robustness performance. }
    \label{fig:rc-ex}
\end{figure}

The error retention curve (RCs) \citep{malinin2019uncertainty,shifts20, brats} assess the correspondence between a chosen uncertainty measure and an error or a quality metric. By quantifying this correspondence for various uncertainty measures we can choose a measure that is better at pointing out errors in model predictions. This is relevant for clinical applications, where uncertainty constitutes a signaling system requiring human verification.
 
Compared to the uncertainty calibration analysis \citep{uncs_survey}, error RCs only consider the ranking of uncertainty values within a particular scan, thus, avoiding uncertainty values scaling present in the calibration metrics. Additionally, they allow for the choice of a quality metric w.r.t. to which the uncertainty measure is compared. Thus, allowing for extending their definition to different scales, \textit{e.g.} lesion or patient. Moreover, compared to calibration metrics, the RC analysis allows us to estimate the upper and lower bounds of the uncertainty-robustness performance.

\paragraph{Voxel-scale DSC-RC} Similarly to our previous investigation \citep{molchanova_isbi}, we use voxel-scale RCs to quantify the average across patients correspondence between per-voxel uncertainty and DSC, \textit{i.e.} per-voxel misclassification errors of different kinds: either FP or FN. For one patient, a voxel-scale DSC-RC is built by sequentially replacing a fraction \( \tau \) of the most uncertain voxel predictions within the brain mask with the ground truth and re-computing the DSC. If one measure has a better ability to capture model errors than another measure, then the most uncertain voxels will be faster replaced with the ground truth and the DSC-RC will grow faster. Thus, the area under the DSC retention curve (DSC-RC), further referred to as DSC-AUC can be used to compare different uncertainty measures in their ability to capture model segmentation errors. It is possible to estimate lower and upper bounds of performance by building \emph{random} and \emph{ideal} RCs. For a random RC, we assign random uncertainty values to each voxel of predictions. For the ideal one, a zero uncertainty is assigned to true positive and negative (TP and TN) voxels while false positive and negative (FP and FN) voxels have an uncertainty of 1. To build the RCs, we use $\tau=2.5\cdot10^{-3}$. An illustrative explanation of a voxel-scale RC can be found in Figure \ref{fig:rc-ex}. 

\paragraph{Lesion-scale LPPV-RC (proposed)} In our previous investigation \citep{molchanova_isbi} we proposed an extension of the error RC analysis to the lesion scale through LF1-RC. LF1-RC assesses the correspondence between lesion-scale uncertainty and errors in lesion detection within a patient. As defined in Section \ref{sec:error}, the LF1 is reflective of both FNL and FPL. However, uncertainty cannot be defined for FNLs as they are not predicted, but ground-truth lesions. Thus, LF1-RCs are more suitable for the comparison of different models or uncertainty quantification methods, for which the number of FNL can vary. However, for the comparison of lesion-scale uncertainty measures, where the number of FNLs does not change, the LPPV-RC analysis is sufficient. Thus, we propose the LPPV-RC assesses the correspondence between lesion-scale uncertainty and lesion false positive errors within a patient. Intuitively, this analysis helps to understand which uncertainty measure is the best at pointing to false positive lesions.

Building a LPPV-RC for a patient starts with computing the number of TPL and FPL, \textit{i.e.} $\#_{TPL}$ and $\#_{FPL}$, and uncertainty values for each of these lesions. Further, the most uncertain lesions are sequentially replaced with TPL, and LPPV is recomputed. Analogously to the voxel scale, if a lesion-scale uncertainty measure has a better ability to capture FPL than another measure, then FPL will be replaced faster, and the curve will grow faster. Thus, the area under the LPPV-RC, that is LPPV-AUC, can be used to compare different measures in their ability to capture FPL detection errors. As each patient has a different number of predicted lesions, to obtain an average across the dataset LPPV-AUC, we first need to interpolate all LPPV-RCs to a similar set of retention fractions. For this, we use a piecewise linear interpolation and a set of retention fractions similar to the voxel scale. Additionally, similarly to the voxel scale, the ideal and random RCs are built. The ideal curve is built by considering all TPLs having an uncertainty of 0 and all FPLs having an uncertainty of 1. The random curve is built by using random uncertainties for each of the lesions.

\paragraph{Patient-scale DSC-RC (proposed)} In this work, we propose a way to extend an error RCs analysis to the patient scale to assess the correspondence between patient-scale uncertainty measures and overall prediction quality in a patient. We use DSC as a measure of overall segmentation quality. Then, a patient-scale DSC-RC is built by sequentially excluding the most uncertain patients, that is replacing their DSC with 1.0, and recomputing the average across the dataset DSC. Similarly to the voxel and lesion scales, the area under the patient-scale DSC-RC is used to compare the ability of different patient-scale uncertainty measures to capture patients with a greater number of erroneous predictions. In analogy to the voxel and lesion scales, we want to assess the upper and lower bounds of the performance with ideal and random patient-scale DSC-RCs. To build a random curve we assign random uncertainties to each of the patients. To build the ideal curve, we use a negated DSC score as an uncertainty measure, as we want ideal uncertainty to point to the most erroneous examples in terms of DSC.

\paragraph{Statistical testing} For the voxel and lesion scales, the error retention curves analysis, namely DSC-RC and LPPV-RC, are computed per patient. Therefore, when comparing different uncertainty measures across each other, one can assess the differences in AUC distributions across measures, \textit{e.g.} statistics. For the patient scale, DSC-RC is computed per dataset (by iterative replacement of the most uncertain patients). Nevertheless, it is possible to estimate the bootstrap confidence intervals by treating the patient-scale DSC-RC as a statistic itself. Thus, to conduct the measures ranking for the patient-scale uncertainty measures, we compare the mean patient-scale DSC-AUC, paying attention to the corresponding confidence intervals.

\subsubsection{Patient-scale uncertainty as a proxy for segmentation quality}

In addition to the information brought by the error RC, we would like to study if a patient-scale uncertainty can serve as a proxy to the model segmentation quality, measured by DSC. For this, we compute Spearman's correlation coefficient $\rho$ between the DSC and uncertainty values. The Spearman’s correlation is computed for different test sets separately, and then jointly. The joint correlation coefficient should show if the uncertainty measure can be used as a proxy for the segmentation quality regardless of the domain shift. This might be particularly useful for the scenario where the domain shift is unknown.

\subsection{WML segmentaiton model}

For this study, we consider two models based on a 3D U-Net architecture. Similar 3D-U-net-based models have been previously used for WML segmentation and compared to other approaches \citep{sotamsseg, mssegreview, LAROSA2020102335, SPAGNOLO2023103491}. Furthermore, our choice is supported by the fact that the same model has been extensively used previously for UQ exploration within the same WML segmentation task in MS \citep{MCKINLEY2020102104, TAlber, shifts20, Dojat, DojatMICCAI}. The first model is the baseline model from the Shifts 2.0 Challenge \citep{shifts20} dedicated to UQ for WML segmentation. The second model is a self-configuring nnU-Net architecture \citep{isensee2019}. Both models are ensembles with 5 members, where each member is a 3D U-Net model \citep{ronneberger-2015, cicek-2016}. There are several crucial differences between the Shifts Baseline (SB) U-Net and the nnU-Net models: i) architecture, i.e SB has the depth reduced by one and, thus, less trainable parameters; ii) loss function, \textit{i.e.} Focal-Dice loss for SB and cross-entropy and Dice loss for nnU-Net; iii) deep supervision is utilized by nnU-Net, compared to SB; iv) input, SB’s input are patches of the size  $96\times96\times96$ cropped from the brain using a sequence of transforms, while nnU-Net uses patches $112\times160\times128$ cropped around the whole brain. Both models represent public benchmarks, and their training and inference code is available online~\footnote{The original code including model implementation and weights, training and inference code can be found at the Shifts Challenge GitGub: \url{https://github.com/Shifts-Project/shifts/tree/main/mswml}. nnU-Net model code is publicly available at \url{https://github.com/MIC-DKFZ/nnUNet}. Model weights can be found on our GitHub: \url{https://github.com/Medical-Image-Analysis-Laboratory/MS_WML_uncs}.}.  For the SB model, the only difference, compared to the original model, is an addition of 2 more ensemble members, obtained using the original training code. For the nnU-Net model, we used a “3d\_fullres” configuration, we ensured the consistency of training and validation examples across folds (for the model to be comparable to SB) and limited the number of training epochs to 200 (due to the validation loss stagnation, to prevent overfitting). Since the Shifts dataset does not contain lesions less than 10 voxels, we process the outputs of each of the models to remove all the connected components with less than 10 voxels.

%% file: results.tex
\section{Results}

\subsection{Model performance}

The evaluation of the ensemble model performance in terms of average segmentation and lesion detection quality is presented in Table \ref{tab:perf} for training, validation, and testing sets. Regardless of the model, SB or nnU-Net, the in-domain performance reaches its upper bound determined by the inter-rater agreement reported in. There is a considerable drop in performance (around 10\% depending on the metric) between in- and out-of-domain sets both in terms of segmentation (DSC and nDSC) and lesion detection (LF1). The performance on Test$_{private}$ and Test$_{WMH}$ datasets lies in between Test$_{in}$ and Test$_{out}$ with regards to segmentation and lesion detection quality.   Between the two models, nnU-Net shows higher performance in terms of segmentation and lesion detection.

\begin{table}[htbp]
\centering
\tiny
\begin{tabular}{|c|cc|cc|cc|cc|}
\hline
\textbf{Set}     & \multicolumn{2}{c|}{\textbf{DSC}} & \multicolumn{2}{c|}{\textbf{nDSC}} & \multicolumn{2}{c|}{\textbf{LF1}} & \multicolumn{2}{c|}{\textbf{LPPV}} \\ \hline
                 & \textbf{SB}   & \textbf{nnU-Net}   & \textbf{SB}   & \textbf{nnU-Net}  & \textbf{SB}  & \textbf{nnU-Net}  & \textbf{SB}   & \textbf{nnU-Net} \\ \hline
Train   & \parbox{1cm}{0.756 [0.737, 0.774]} & \parbox{1cm}{0.906 [0.892, 0.917]} & \parbox{1cm}{0.725 [0.699, 0.749]} & \parbox{1cm}{0.856 [0.826, 0.883]} & \parbox{1cm}{0.547 [0.493, 0.596]} & \parbox{1cm}{0.845 [0.787, 0.876]} & \parbox{1cm}{0.689 [0.627, 0.735]} & \parbox{1cm}{0.971 [0.957, 0.981]} \\ \hline
Val     & \parbox{1cm}{0.720 [0.602, 0.783]} & \parbox{1cm}{0.776 [0.701, 0.821]} & \parbox{1cm}{0.684 [0.625, 0.740]} & \parbox{1cm}{0.736 [0.669, 0.783]} & \parbox{1cm}{0.444 [0.345, 0.547]} & \parbox{1cm}{0.643 [0.555, 0.707]} & \parbox{1cm}{0.533 [0.425, 0.608]} & \parbox{1cm}{0.762 [0.624, 0.871]} \\ \hline
Test$_{in}$    & \parbox{1cm}{0.633 [0.582, 0.673]} & \parbox{1cm}{0.707 [0.671, 0.739]} & \parbox{1cm}{0.689 [0.662, 0.717]} & \parbox{1cm}{0.741 [0.715, 0.768]} & \parbox{1cm}{0.487 [0.439, 0.528]} & \parbox{1cm}{0.701 [0.666, 0.733]} & \parbox{1cm}{0.610 [0.552, 0.660]} & \parbox{1cm}{0.762 [0.721, 0.797]} \\ \hline
Test$_{out}$   & \parbox{1cm}{0.488 [0.457, 0.515]} & \parbox{1cm}{0.571 [0.538, 0.600]} & \parbox{1cm}{0.533 [0.501, 0.560]} & \parbox{1cm}{0.603 [0.570, 0.630]} & \parbox{1cm}{0.333 [0.308, 0.361]} & \parbox{1cm}{0.502 [0.477, 0.525]} & \parbox{1cm}{0.623 [0.586, 0.659]} & \parbox{1cm}{0.828 [0.799, 0.852]} \\ \hline
Test$_{private}$ & \parbox{1cm}{0.601 [0.578, 0.621]} & \parbox{1cm}{0.646 [0.626, 0.665]} & \parbox{1cm}{0.628 [0.608, 0.645]} & \parbox{1cm}{0.653 [0.635, 0.670]} & \parbox{1cm}{0.416 [0.396, 0.437]} & \parbox{1cm}{0.562 [0.543, 0.581]} & \parbox{1cm}{0.581 [0.556, 0.605]} & \parbox{1cm}{0.799 [0.779, 0.817]} \\ \hline
Test$_{WMH}$    & \parbox{1cm}{0.591 [0.564, 0.616]} & \parbox{1cm}{0.648 [0.623, 0.671]} & \parbox{1cm}{0.599 [0.580, 0.617]} & \parbox{1cm}{0.651 [0.632, 0.668]} & \parbox{1cm}{0.373 [0.353, 0.391]} & \parbox{1cm}{0.555 [0.534, 0.574]} & \parbox{1cm}{0.488 [0.456, 0.518]} & \parbox{1cm}{0.696 [0.665, 0.724]} \\ \hline
\end{tabular}
\caption{Mean average model performance in segmentation (DSC and nDSC) and lesion detection (LF1 and LPPV). 90\% confidence intervals were computed using bootstrapping. SB - Shifts 2.0 Challenge baseline model.}
\label{tab:perf}
\end{table}

\subsection{Quantitative evaluation of uncertainty measures}

\subsubsection{Error retention curve analysis}

\begin{figure}[h!]
    \centering
    \includegraphics[width=\linewidth]{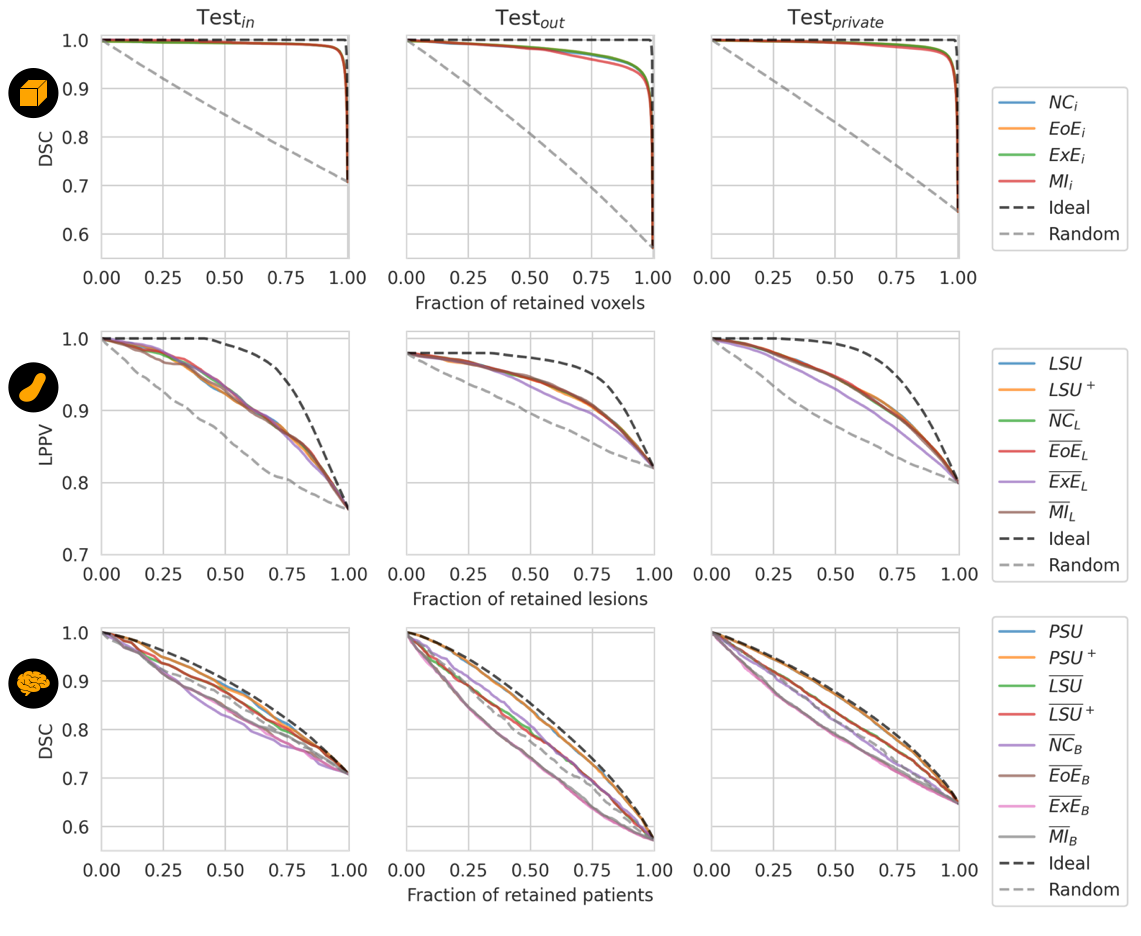}
    \caption{Error retention curves for the assessment of uncertainty measures at the voxel, lesion, and patient anatomical scales across the in-domain Test$_{in}$ (left column) and the out-of-domain Test$_{out}$ (center column) and Test$_{private}$ (left column) sets for the nnU-Net model. Different rows correspond to different anatomical scales indicated with icons on the left. The voxel-scale DSC-RCs and lesion-scale LPPV-RCs were obtained by averaging across the respective datasets. At each of the scales, the ideal (black dashed) line indicates the upper bound of an uncertainty measure performance in its ability to capture model errors; the random (gray dashed) indicates no relationship between an uncertainty measure and error; a worse-than-random performance indicates an inverse relationship. Analogous results for the SB model are shown in \ref{appendix:erc-sb}. }
    \label{fig:erc}
\end{figure}

The RCs for the assessment of uncertainty measures on each of the anatomical scales (voxel, lesion, and patient) are presented in Figure \ref{fig:erc}. The voxel-scale DSC-RCs and lesion-scale LPPV-RCs were obtained by averaging across the respective datasets. The mean areas under the error retention curves and the results of the statistical testing are presented in Table \ref{tab:erc}. 

Regardless of the test set, all \textbf{voxel-scale} uncertainty measures outperform random uncertainty and are closer to the ideal uncertainty in terms of mean DSC-AUC, indicating their ability to capture errors in segmentation. However, the marginal difference between DSC-AUCs of different measures is relatively small. On the in-domain Test$_{in}$, there is no agreement between two models in terms of the measures with the highest mean DSC-AUC: while total and data uncertainty ($NC_i, EoE_i, ExE_i$) have higher DSC-AUC for the SB model, model uncertainty ($MI_i$) has a higher DSC-AUC for the nnU-Net model. On the out-of-domain Test$_{out}$ and Test$_{private}$ datasets, the entropy-based total and data uncertainty measures ($EoE_i$ and $ExE_i$) tend to have an advantage compared to other measures, contributing to their overall advantage in the whole evaluation. Nevertheless, the aggregation of data uncertainty $ExE_i$ for the lesion-/ patient- uncertainty computation usually yields the worst results in terms of lesion-scale LPPV-AUC / patient-scale DSC-AUC. This means that a good performance of an uncertainty measure in capturing voxel misclassifications, when aggregated, does not necessarily lead to an optimal uncertainty measure for detecting lesion false positive or overall segmentation failure. 

Regardless of the test set, at the \textbf{lesion scale}, there is a greater marginal difference between different measures, particularly for the SB model. For the SB model, the proposed measure $LSU^+$ has an advantage in the mean LPPV-AUC over other measures, indicating a better ability to capture lesion false positive errors. While $LSU$ and $LSU^+$ have similar LPPV-AUCs, there is usually some difference in their performances, benefiting the $LSU^+$ measure. Among the measures based on the aggregation of voxel uncertainties, aggregated total uncertainty $\overline{EoE}_{L},$ generally provides slightly higher mean LPPV-AUC. Despite the differences between the mean LPPV-AUCs among lesion-scale measures, the 90\% confidence interval overlap suggests that these differences are limited.

At the \textbf{patient scale}, the marginal differences between various measures are prominent compared to the voxel and lesion scales, especially on the out-of-domain sets. The results are aligned for both in- and out-of-domain test sets and for both models, SB and nnU-Net. The proposed $PSU$ and $PSU^+$  measures have comparable and the highest patient-scale DSC-AUCs, suggesting their superior ability to capture overall segmentation failure. The aggregation of the best in terms of LPPV-AUC lesion scale uncertainty (\textit{i.e.} $LSU$ and $LSU^+$) yields lower patient DSC-AUC. Averaging voxel uncertainties across the brain generally provides worse-than-random performance in the error retention curve analysis. The last means that an average across-subject voxel-scale uncertainty is not informative of an overall segmentation performance on a particular subject measured by DSC or has an inverse relationship with errors.

\begin{table}[h!]
\tiny
\centering
\begin{tabular}{|c|cc|cc|cc|}
\hline
\textbf{Measure} & \multicolumn{2}{c|}{\textbf{Test$_{in}$}} & \multicolumn{2}{c|}{\textbf{Test$_{out}$}} & \multicolumn{2}{c|}{\textbf{Test$_{private}$}} \\ \hline
                 & \textbf{SB}   & \textbf{nnU-Net}   & \textbf{SB}   & \textbf{nnU-Net}  & \textbf{SB}  & \textbf{nnU-Net} \\ \hline
                 
\multicolumn{7}{|c|}{\textbf{Voxel-scale DSC-AUC ($\uparrow$)}} \\ \hline

\textbf{Ideal}   & \parbox{1.5cm}{99.93 [99.91, 99.94]} & \parbox{1.5cm}{99.94 [99.92, 99.95]} & \parbox{1.5cm}{99.90 [99.88, 99.91]} & \parbox{1.5cm}{99.93 [99.92, 99.92]} & \parbox{1.5cm}{99.93 [99.91, 99.94]} & \parbox{1.5cm}{99.93 [99.91, 99.94]} \\
\textbf{$NC_i$}   & \parbox{1.5cm}{\textbf{99.17} [98.99, 99.31]} & \parbox{1.5cm}{99.17 [98.29, 99.49]} & \parbox{1.5cm}{96.74 [96.23, 97.12]} & \parbox{1.5cm}{97.59 [97.02, 0.9797]} & \parbox{1.5cm}{98.56 [98.36, 98.70]} & \parbox{1.5cm}{\textbf{99.02} [98.82, 99.16]} \\
\textbf{$EoE_i$}   & \parbox{1.5cm}{\textbf{99.16} [98.99, 99.31]} & \parbox{1.5cm}{\textit{99.11} [98.10, 99.46]} & \parbox{1.5cm}{\textbf{97.02} [96.56, 97.37]} & \parbox{1.5cm}{\textbf{97.72} [97.22, 98.05]} & \parbox{1.5cm}{\textbf{98.65} [98.46, 98.79]} & \parbox{1.5cm}{\textbf{99.02} [98.82, 99.17]} \\
\textbf{$ExE_i$}   & \parbox{1.5cm}{\textbf{99.16} [98.99, 99.31]} & \parbox{1.5cm}{\textit{99.11} [98.09, 99.46]} & \parbox{1.5cm}{\textbf{97.02} [96.56, 97.38]} & \parbox{1.5cm}{\textbf{99.71} [97.21, 98.05]} & \parbox{1.5cm}{\textbf{98.65} [98.46, 98.80]} & \parbox{1.5cm}{\textbf{99.02} [98.82, 99.16]} \\
\textbf{$MI_i$}   & \parbox{1.5cm}{\textit{99.05} [98.85, 99.21]} & \parbox{1.5cm}{\textbf{99.27} [98.74, 99.50]} & \parbox{1.5cm}{\textit{96.69} [96.19, 97.08]} & \parbox{1.5cm}{\textit{97.28} [96.70, 97.68]} & \parbox{1.5cm}{\textit{98.46} [98.25, 98.62]} & \parbox{1.5cm}{\textit{98.86} [98.63, 99.01]} \\
\textbf{Random}  & \parbox{1.5cm}{80.91 [76.77, 83.36]} & \parbox{1.5cm}{84.87 [82.79, 86.69]} & \parbox{1.5cm}{76.20 [74.88, 77.36]} & \parbox{1.5cm}{80.00 [78.72, 81.21]} & \parbox{1.5cm}{80.18 [78.99, 81.19]} & \parbox{1.5cm}{82.79 [81.85, 83.62]} \\ \hline

\multicolumn{7}{|c|}{\textbf{Lesion-scale LPPV-AUC ($\uparrow$)}} \\ \hline

\textbf{Ideal}   & \parbox{1.5cm}{87.88 [82.60, 90.91]} & \parbox{1.5cm}{95.72 [93.89, 96.88]} & \parbox{1.5cm}{87.07 [83.40, 89.46]} & \parbox{1.5cm}{96.47 [93.13, 97.66]} & \parbox{1.5cm}{86.41 [84.54, 87.93]} & \parbox{1.5cm}{96.36 [95.51, 96.96]} \\
\textbf{$LSU$}   & \parbox{1.5cm}{83.54 [75.80, 87.04]} & \parbox{1.5cm}{91.54 [89.57, 93.15]} & \parbox{1.5cm}{83.28 [79.63, 85.91]} & \parbox{1.5cm}{\textbf{94.06} [90.87, 95.41]} & \parbox{1.5cm}{82.63 [80.74, 84.28]} & \parbox{1.5cm}{\textbf{93.29} [92.15, 94.21]} \\
\textbf{$LSU^+$}   & \parbox{1.5cm}{\textbf{83.90} [78.83, 87.31]} & \parbox{1.5cm}{91.51 [89.53, 93.12]} & \parbox{1.5cm}{\textbf{83.89} [80.27, 86.45]} & \parbox{1.5cm}{93.97 [90.80, 95.33]} & \parbox{1.5cm}{\textbf{82.70} [80.83, 84.37]} & \parbox{1.5cm}{\textbf{93.29} [92.15, 94.20]} \\
\textbf{$\overline{NC}_{L}$}   & \parbox{1.5cm}{83.33 [78.34, 86.77]} & \parbox{1.5cm}{91.71 [89.46, 93.92]} & \parbox{1.5cm}{83.24 [79.60, 85.86]} & \parbox{1.5cm}{\textbf{94.06} [90.84, 95.39]} & \parbox{1.5cm}{82.34 [80.38, 84.04]} & \parbox{1.5cm}{93.14 [92.05, 94.05]} \\
\textbf{$\overline{EoE}_{L}$}   & \parbox{1.5cm}{83.38 [78.41, 86.83]} & \parbox{1.5cm}{\textbf{91.81} [89.61, 93.93]} & \parbox{1.5cm}{83.26 [79.63, 85.88]} & \parbox{1.5cm}{\textbf{94.07} [90.86, 95.40]} & \parbox{1.5cm}{82.28 [80.30, 83.99]} & \parbox{1.5cm}{93.22 [92.11, 94.11]} \\
\textbf{$\overline{ExE}_{L}$}   & \parbox{1.5cm}{\textit{81.73} [76.70, 85.24]} & \parbox{1.5cm}{91.70 [89.50, 93.27]} & \parbox{1.5cm}{\textit{81.55} [77.88, 84.17]} & \parbox{1.5cm}{\textit{93.41} [90.32, 94.77]} & \parbox{1.5cm}{\textit{78.74} [76.80, 80.56]} & \parbox{1.5cm}{\textit{91.99} [90.77, 93.00]} \\
\textbf{$\overline{MI}_{L}$}   & \parbox{1.5cm}{82.63 [77.70, 86.03]} & \parbox{1.5cm}{\textit{91.37} [89.22, 92.98]} & \parbox{1.5cm}{82.31 [78.64, 85.00]} & \parbox{1.5cm}{\textbf{94.06} [90.86, 95.40]} & \parbox{1.5cm}{81.62 [79.69, 83.34]} & \parbox{1.5cm}{93.05 [91.89, 93.96]} \\
\textbf{Random}  & \parbox{1.5cm}{76.69 [71.57, 80.48]} & \parbox{1.5cm}{86.65 [83.96, 88.94]} & \parbox{1.5cm}{76.35 [72.71, 79.19]} & \parbox{1.5cm}{90.59 [87.65, 92.10]} & \parbox{1.5cm}{73.97 [71.91, 75.81]} & \parbox{1.5cm}{88.61 [87.18, 89.88]} \\ \hline

\multicolumn{7}{|c|}{\textbf{Patient-scale DSC-AUC ($\uparrow$)}} \\ \hline

\textbf{Ideal}   & \parbox{1.5cm}{85.74 [84.16, 87.52]} & \parbox{1.5cm}{88.72 [87.22, 90.36]} & \parbox{1.5cm}{79.21 [77.96, 80.52]} & \parbox{1.5cm}{83.55 [82.30, 84.95]} & \parbox{1.5cm}{84.48 [83.72, 85.26]} & \parbox{1.5cm}{86.23 [85.56, 86.91]} \\
\textbf{$PSU$}   & \parbox{1.5cm}{\textbf{84.99} [83.16, 86.81]} & \parbox{1.5cm}{\textbf{87.90} [86.25, 89.73]} & \parbox{1.5cm}{\textbf{78.40} [77.11, 79.73]} & \parbox{1.5cm}{\textbf{82.68} [81.26, 84.18]} & \parbox{1.5cm}{\textbf{83.63} [82.79, 84.47]} & \parbox{1.5cm}{\textbf{85.73} [85.02, 86.46]} \\
\textbf{$PSU^+$}   & \parbox{1.5cm}{\textbf{84.82} [82.97, 86.68]} & \parbox{1.5cm}{\textbf{87.84} [86.17, 89.70]} & \parbox{1.5cm}{\textbf{78.39} [77.10, 79.70]} & \parbox{1.5cm}{\textbf{82.70} [81.28, 84.20]} & \parbox{1.5cm}{\textbf{83.60} [82.75, 84.44]} & \parbox{1.5cm}{\textbf{85.75} [85.04, 86.47]} \\
\textbf{$\overline{LSU}$}   & \parbox{1.5cm}{83.77 [81.99, 85.42]} & \parbox{1.5cm}{86.80 [84.69, 88.64]} & \parbox{1.5cm}{75.48 [74.55, 77.26]} & \parbox{1.5cm}{79.90 [77.66, 81.22]} & \parbox{1.5cm}{79.91 [80.02, 82.28]} & \parbox{1.5cm}{83.55 [82.54, 84.44]} \\
\textbf{$\overline{LSU^+}$}   & \parbox{1.5cm}{83.13 [81.04, 84.88]} & \parbox{1.5cm}{86.87 [84.80, 88.97]} & \parbox{1.5cm}{75.28 [74.36, 77.08]} & \parbox{1.5cm}{79.76 [75.73, 81.16]} & \parbox{1.5cm}{79.91 [80.02, 82.28]} & \parbox{1.5cm}{83.52 [82.51, 84.42]} \\
\textbf{$\overline{NC}_{B}$}   & \parbox{1.5cm}{80.70 [78.27, 82.42]} & \parbox{1.5cm}{\textit{84.13} [82.40, 85.57]} & \parbox{1.5cm}{74.82 [72.96, 76.57]} & \parbox{1.5cm}{79.90 [77.85, 81.73]} & \parbox{1.5cm}{79.79 [75.81, 78.79]} & \parbox{1.5cm}{82.00 [80.84, 82.97]} \\
\textbf{$\overline{EoE}_{B}$}   & \parbox{1.5cm}{80.19 [76.20, 82.86]} & \parbox{1.5cm}{84.71 [82.40, 86.80]} & \parbox{1.5cm}{\textit{71.60} [69.62, 73.32]} & \parbox{1.5cm}{75.04 [72.69, 76.72]} & \parbox{1.5cm}{77.43 [75.81, 78.79]} & \parbox{1.5cm}{79.98 [78.49, 81.19]} \\
\textbf{$\overline{ExE}_{B}$}   & \parbox{1.5cm}{\textit{80.19} [76.20, 82.87]} & \parbox{1.5cm}{84.64 [82.34, 86.72]} & \parbox{1.5cm}{\textit{71.57} [69.53, 73.31]} & \parbox{1.5cm}{\textit{75.44} [72.69, 76.72]} & \parbox{1.5cm}{\textit{77.37} [75.74, 78.73]} & \parbox{1.5cm}{\textit{79.83} [78.33, 81.04]} \\
\textbf{$\overline{MI}_{B}$}   & \parbox{1.5cm}{80.28 [76.25, 83.02]} & \parbox{1.5cm}{85.07 [82.69, 87.20]} & \parbox{1.5cm}{71.70 [69.76, 73.39]} & \parbox{1.5cm}{75.18 [72.93, 77.00]} & \parbox{1.5cm}{77.60 [75.97, 78.97]} & \parbox{1.5cm}{80.20 [78.72, 81.42]} \\
\textbf{Random}  & \parbox{1.5cm}{81.87 [79.09, 83.84]} & \parbox{1.5cm}{85.73 [83.58, 87.62]} & \parbox{1.5cm}{74.10 [72.53, 75.53]} & \parbox{1.5cm}{78.03 [76.20, 79.60]} & \parbox{1.5cm}{80.08 [78.85, 81.14]} & \parbox{1.5cm}{82.26 [81.14, 83.23]} \\ \hline

\end{tabular}
\caption{Mean average areas under error retention curves and 90\% bootstrap confidence intervals for the assessment of the uncertainty measures at the voxel, lesion, and patient anatomical scales across the in-domain Test$_{in}$ (left column) and the out-of-domain Test$_{out}$ (center column) and Test$_{private}$ (right column) sets. Results are presented for the Shifts Challenge Baseline (SB) and nnU-Net models. Highest AUC values for each dataset, model, and anatomical scale are highlighted in \textbf{bold}, lowest - in \textit{italic}; ideal and random values are in gray colour and indicate the upper and lower bounds of performance, respectively.}
\label{tab:erc}
\end{table}

\subsubsection{Patient-scale uncertainty as a proxy to the segmentation quality}

\begin{figure}[h!]
    \centering
    \includegraphics[width=\linewidth]{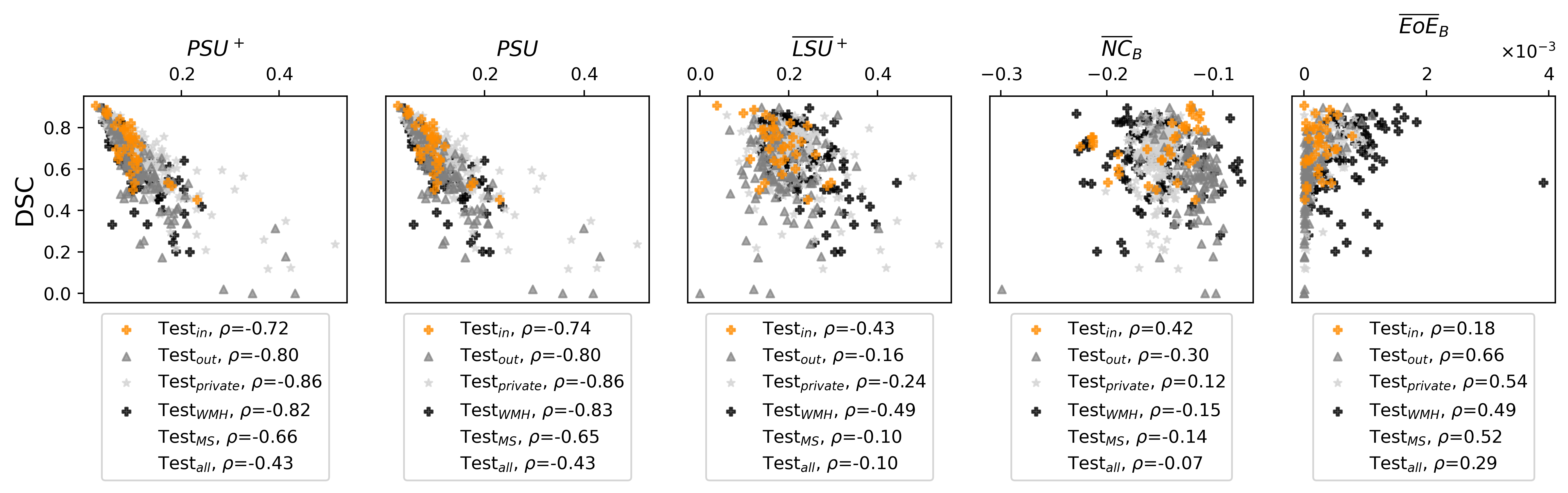}
    \caption{The relationship between DSC and patient-scale uncertainty is assessed for Test$_{in}$ (orange), Test$_{out}$ (gray), Test$_{private}$ (light gray), and Test$_{WMH}$ (black) separately and jointly for the nnU-Net model. The presented uncertainty measures were chosen based on the results of the error RC analysis (Figure \ref{fig:erc} and Table \ref{tab:erc}) to illustrate the relationship between DSC and uncertainty brought by measures with the highest (proposed $PSU^{(+)}$), median (proposed $\overline{LSU^{(+)}}$), and worse-than-random ($\overline{NC}_B$ and $\overline{EoE}_B$) DSC-AUC values. Results for other measures and for the SB model can be found in \ref{appendices:corr-pat}.}
    \label{fig:corr}
\end{figure}

Extending the analysis of the relationship between the patient-scale uncertainty measures and the segmentation quality measures by DSC, Table \ref{tab:corr} presents corresponding Spearman's correlation coefficients. Figure \ref{fig:corr} contains plots DSC and patient uncertainty for the measures with the highest (proposed $PSU^{(+)}$), median (proposed $\overline{LSU^{(+)}}$), and worse-than-random ( $\overline{NC}_B$ and $\overline{EoE}_B$) patient-scale DSC-AUC values. For the SB model and the rest of the measures, the same analysis and trends can be found in \ref{appendices:corr-pat}. The results show the highest correlation between the patient uncertainty and DSC is provided by the proposed $PSU^{(+)}$ measures, with $\rho$ around 0.8 across different test sets. For the aggregation of the lesion-scale uncertainty, the correlation with the segmentation quality drops at least twice. For the measures based on the voxel-scale uncertainty aggregation, the correlation is either weak, \textit{e.g.} $\overline{NC}_B$, or positive. There is a positive correlation between $\overline{EoE}_B$, $\overline{ExE}_B$, and $\overline{MI}_B$, suggesting that high uncertainty can point to examples with high DSC. The absolute value of this correlation is around 0.5, which is higher than for $\overline{LSU^{(+)}}$, yet lower than for the proposed $PSU^{(+)}$.

\begin{table}[h!]
\small
\centering
\begin{tabular}{|c|cc|cc|cc|cc|cc|}
\hline
\textbf{Measures} & \multicolumn{2}{c|}{Test$_{in}$} & \multicolumn{2}{c|}{Test$_{out}$} & \multicolumn{2}{c|}{Test$_{private}$} & \multicolumn{2}{c|}{Test$_{WMH}$} \\\hline
& \textbf{SB}   & \textbf{nnU-Net}   & \textbf{SB}   & \textbf{nnU-Net}  & \textbf{SB}  & \textbf{nnU-Net}  & \textbf{SB}   & \textbf{nnU-Net} \\ \hline
$PSU$      & \textbf{-0.81}  & \textbf{-0.74}  & \textbf{-0.81}  & \textbf{-0.84}  & \textbf{-0.86}  & \textbf{-0.87}  & \textbf{-0.83}  & \textbf{-0.46} \\
$PSU^+$      & -0.72  & -0.72  & -0.80  & \textbf{-0.84}  & -0.86  & -0.86  & \textbf{-0.83}  & -0.43 \\
$\overline{LSU}$      & -0.41  & -0.41  & -0.22  & -0.37  & -0.25  & -0.36  & -0.49  & -0.11 \\
$\overline{LSU}^+$     & -0.29  & -0.43  & -0.22  & -0.34  & -0.42  & -0.49  & -0.11  & -0.11 \\
$\overline{NC}_B$      & 0.36   & 0.42   & 0.11   & 0.30   & 0.30   & 0.14   & -0.09  & 0.07  \\
$\overline{EoE}_B$       & 0.23   & 0.18   & 0.55   & 0.56   & 0.54   & 0.54   & 0.53   & 0.31  \\
$\overline{ExE}_B$      & 0.23   & 0.21   & 0.55   & 0.68   & 0.56   & 0.57   & 0.57   & 0.31  \\
$\overline{MI}_B$      & 0.20   & 0.07   & 0.53   & 0.63   & 0.49   & 0.47   & 0.33   & 0.24  \\ \hline
\end{tabular}
\caption{Spearman's correlation coefficients quantifying the relationship between different patient-scale uncertainty values and segmentation quality measured by DSC for different test sets and their combinations. The highest negative correlation values are highlighted in \textbf{bold}.}
\label{tab:corr}
\end{table}

\pagebreak

\subsubsection{Generaliability of the analysis on white matter hyperintensity}

Beyond MS patients, the multi-scale error retention curve and DSC-uncertainty correlation analyses were replicated on a large publicly available cohort of subjects with WMH (Test$_{WMH}$). The full analysis is available in \ref{appendix:generalizability}.

The observed performance of the proposed patient-scale measures discussed in the previous sections is replicated for this new task of WMH segmentation. The results in Figure \ref{fig:corr} and Table \ref{tab:corr} confirm that the proposed measures $PSU^{(+)}$ have a stronger relationship segmentation quality compared to the aggregation measures.

\subsection{Qualitative evaluation of the uncertianty maps}

\begin{figure}
    \centering
    \includegraphics[width=\linewidth]{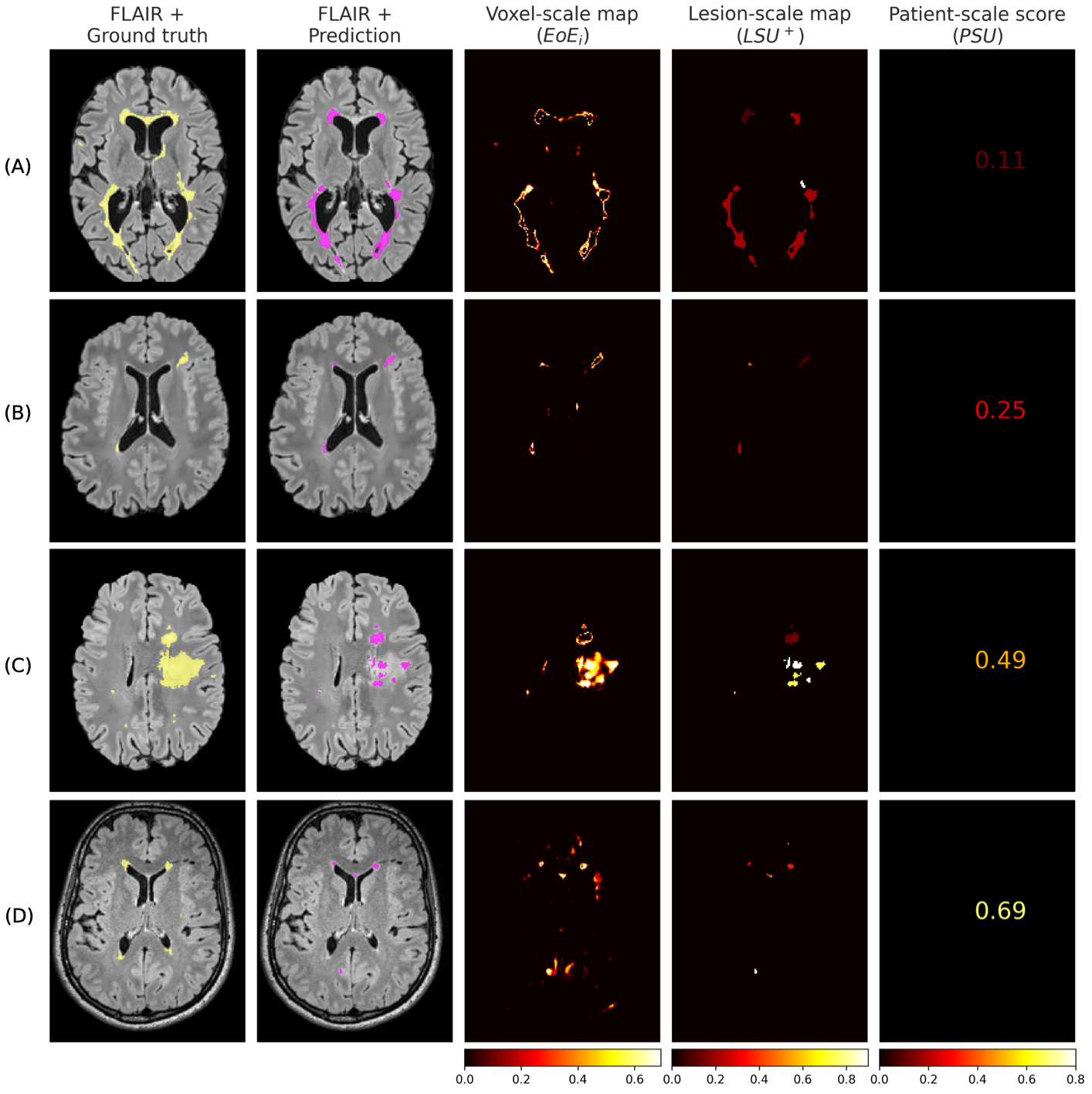}
    \caption{Examples of uncertainty maps at the voxel and lesion scales and patient uncertainty values. The two left columns illustrate axial slices of a FLAIR scan with the ground truth (in yellow) and predicted (in pink) WML masks; the middle column - voxel-scale uncertainty maps computed with the ${EoE}_i$ measure; the fourth column - lesion-scale uncertainty maps computed with the proposed ${LSU}^+$; the fifth column - the patient-scale uncertainty value computed with the proposed ${PSU}^+$. The choice of measures is based on the results of the error retention curves analysis. (A), (B), (C), and (D) represent different scenarios with gradually decreasing DSC. Cases (A) and (B) represent good and mediocre model performance, respectively. Patient (C) has an atypical large lesion, which the algorithm fails as expected. Patient (D) was not correctly preprocessed (the skull is not removed) which led to the algorithm's low performance and high patient uncertainty.}
    \label{fig:qa}
\end{figure}

Our results show that uncertainty quantification mainly at the lesion and patient scales can well depict model error predictions, however, various anatomical scales provide information about different types of errors. In Figure \ref{fig:qa} the uncertainty maps and values are shown for four different subjects, corresponding to different scenarios with respect to the quality of lesion segmentation.

\textbf{Voxel-scale} maps provide refined information about the misclassifications in each voxel. Moreover, voxel-scale uncertainty is always high at the borders of lesions. Hypothetically, this is a reflection of the inter-rater variability or the noise in the ground truth, which are also known to be higher at the borders of lesions. The noise in the data-generation process increases the likelihood of mistakes at the borders of lesions. Nevertheless, the voxel-scale uncertainty can be high in the center of the lesion, signaling that the model is uncertain in the whole lesion region, not only at the borders. Sometimes high uncertainty regions can be related to the FNLs.

\textbf{Lesion-scale} maps provide a visually more intuitive way to assess the correctness of the predicted lesion regions compared to the voxel-scale maps. Particularly, lesion-scale maps can be used to highlight FPLs. Nonetheless, high lesion uncertainty may be an indicator of wrong delineation rather than detection. Let us note that, compared to the voxel-scale, the lesion-scale maps lose all the information about the FNLs.

\textbf{Patient-scale} values inform about the overall quality of the segmentation without indicating the particular reasons for the segmentation failure. As for the chosen examples (C) and (D), high patient uncertainty reveals the fact of the algorithm failure, however for (C) the problem is in the atypical large lesion and for (D) it is a wrong preprocessing, \textit{i.e.} the absence of skull-stripping.

%% file: conclusions.tex
\section{Discussion}

Our research offers a detailed framework for the assessment of uncertainty quantification for a clinically relevant task of white matter lesion segmentation in multiple sclerosis. The specificity of the segmentation task allowed for the exploration of UQ at different anatomical scales: voxel, lesion, and patient. We introduced novel structure-based UQ measures at the lesion and patient scales. For each of these scales, we performed a comparative study between different uncertainty measures (among the state-of-the-art and the proposed) to determine the measures that can point to specific model errors: voxel misclassification, lesion false discovery, or overall low quality of segmentation. For this, we use the error retention curves analysis previously introduced for the pixel or voxel scales \citep{malinin2019uncertainty, shifts20, brats} and extended it to the structural scales in this and our previous work \citep{molchanova_isbi}. Our proposed uncertainty measures ($LSU^{(+)}$ on the lesion scale and $PSU^{(+)}$ on the patient scale from the equations 1-4) quantify the disagreement in the structural predictions between the ensemble model and its members, demonstrating enhanced error detection over state-of-the-art aggregation-based metrics on both in- and out-of-domain datasets. Furthermore, $PSU^{(+)}$ is shown to be a reliable indicator of overall segmentation quality both in- and out-of-domain.

This study compares a variety of voxel-scale measurements adopted from classification tasks, noting their similar capabilities in capturing voxel misclassification errors. A more pronounced difference between these measures is observed after aggregation at other anatomical scales. Particularly, at the lesion scale, higher areas under the respective RCs are observed for the total uncertainty measures, compared to the measure of model uncertainty, and even more data uncertainty. However, voxel uncertainty aggregation at the patient scale yielded results akin to random uncertainty judging by the error RC analysis. Closer examination of the correlation between patient scale uncertainty measures and the DSC revealed a positive relationship, suggesting that a higher average voxel uncertainty correlates with improved DSC.  A high positive correlation of the  aggregation-based measures ( $\overline{EoE}_B$, $\overline{ExE}_B$, and $\overline{MI}_B$) and the total lesion volume in a subject (see \ref{appendix:load}) also goes against common knowledge about the bias in better segmenting subjects with higher lesion loads \citep{Vatsal_isbi}. Similar behavior of the measures based on an aggregation of voxel uncertainties has been previously observed for the task of brain tumor segmentation \citep{10.3389/fnins.2020.00282}, but not for the task of brain structures segmentation \cite{ROY201911}, where the segmented objects are the same and of similar sizes in each of the images. This supports our hypothesis that voxel-scale uncertainty aggregation is unsuitable for tasks affected by this bias. In such cases, structural disagreement metrics present a viable alternative to aggregation-based methods, showing a strong connection to different error types.

\paragraph{Limitations and future work}

The fact that lesion and patient uncertainty measures depend on the choice of the threshold at the model’s output, necessary for the instances or segmented region definition, remains a matter of ongoing debate. We proposed to address the issue by introducing two analogs of the same measure corresponding to different strategies of the threshold choice, \textit{i.e.} $LSU$ versus $LSU^+$ and $PSU$ versus $PSU^+$. Nevertheless, a more detailed investigation of this aspect might be needed. For instance, exploring model calibration as a way to circumvent threshold tuning or investigating measures of uncertainty where this dependence is mitigated.

This paper is focused on the WML segmentation task. While this is a relevant task in clinical practice, there are several medical image segmentation tasks that could adopt the proposed multi-scale approach for UQ. This includes, for instance, nuclei segmentation on histopathology images \citep{kumar-2019}, bone metastases segmentation on the full-body MRI or CT \citep{colombo-2021, afnouch-2023}, vascularized lymph nodes on CT or MRI \citep{hassani2020vascularized}, or white matter lesions in MRI from non-MS patients \citep{malova-2021}. However, finding the multi-center data and benchmarks needed for UQ methods validation under the domain shift in these new tasks remains challenging. 

%% file: acknowledgements.tex
\section*{Acknowledgement}

This work was supported by the Hasler Foundation Responsible AI program (MSxplain), the EU Horizon 2020 project AI4Media (grant 951911), and the Swiss National Science Foundation (SNF) Postdoc Mobility Fellowship (P500PB\_206833). We acknowledge access to the facilities and expertise of the CIBM Center for Biomedical Imaging, a Swiss research center of excellence founded and supported by Lausanne University Hospital (CHUV), University of Lausanne (UNIL), École polytechnique fédérale de Lausanne (EPFL), University of Geneva (UNIGE) and Geneva University Hospitals (HUG). This research is partially funded by the EPSRC (The Engineering and Physical Sciences Research Council) Doctoral Training Partnership (DTP) PhD studentship and supported by Cambridge University Press \& Assessment (CUP\&A), a department of The Chancellor, Masters, and Scholars of the University of Cambridge.

\section*{Declaration of Generative AI Usage and AI-assisted technologies}

During the preparation of this work the authors used ChatGPT-3.5/-4 and Grammarly for the detection of grammatic and stylistic errors in the manuscript. After using these tools, the authors reviewed and edited the content as needed and all take full responsibility for the content of the publication.

%% file: appendix.tex
\section{Definitions of quality metrics}
\label{appendix:metrics}

Let $\#_{TP}, \#_{FP}, \#_{FN}$ be the number of true positive (TP), false positive (FP), and false negative (FN) voxels, respectively.

\textbf{True positive rate:} $$TPR=\frac{\#_{TP}}{\#_{TP} + \#_{FN}}.$$

\textbf{Positive predictive value:} $$PPV = \frac{\#_{TP}}{\#_{TP} + \#_{FP}}.$$

\textbf{Dice similarity score or F$_1$-score}: $$DSC=F_1=\frac{TPR\cdot PPV}{TPR + PPV}=\frac{2 \cdot \#_{TP}}{2 \cdot \#_{TP} + \#_{FP} + \#_{FN}}.$$

\textbf{Normalized Dice similarity score}~\cite{Vatsal_isbi}: $$nDSC=\frac{2 \cdot \#_{TP}}{2 \cdot \#_{TP} + \kappa \cdot \#_{FP} + \#_{FN}}, \kappa = h (r^{-1} - 1).$$

where $h$ represents the ratio between the positive and the negative classes while $0<r<1$ denotes a \emph{reference} value that is set to the mean fraction of the positive class, \textit{i.e.} a lesion class in our case, across a large number of subjects.

Analogous, lesion-scale metrics can be defined by replacing $\#_{TP}$, $\#_{FP}$, $\#_{FN}$ with a number of TP, FP, and FN lesion (TPL, FPL, FNL). As mentioned before, the definition of lesion types can vary. This work uses 25\% overlap to distinguish TPL and FPL among the predicted lesions. FNL is defined as the ground truth lesions that have no overlap with predictions.

\section{Additional results}
\label{appendix:b}

\subsection{Error retention curve analysis for the Shifts 2.0 Challenge Baseline (SB) model}
\label{appendix:erc-sb}

\begin{figure}[h!]
    \centering
    \includegraphics[width=\linewidth]{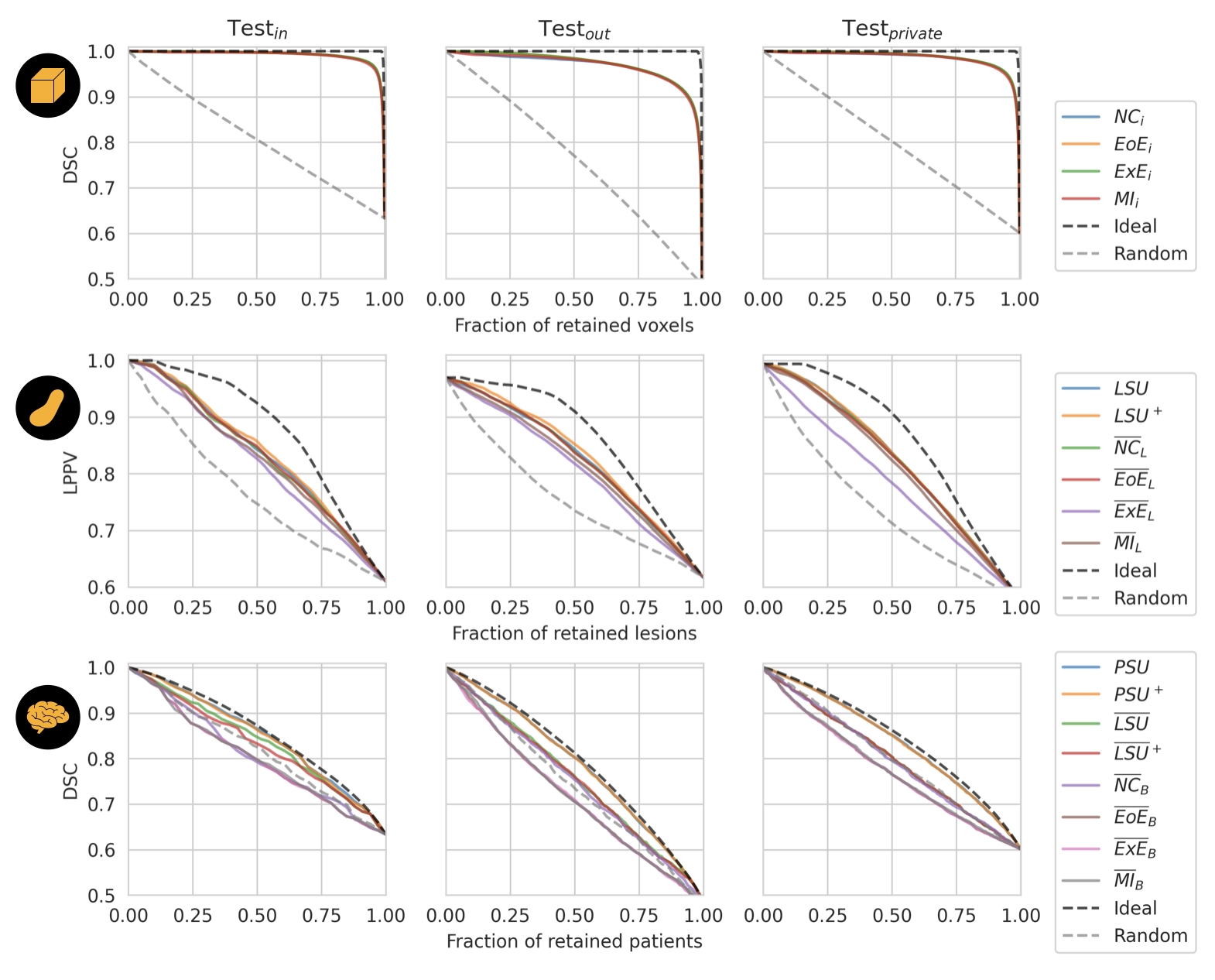}
    \caption{Error retention curves for the assessment of uncertainty measures at the voxel, lesion, and patient (rows one, two, and three, respectively) anatomical scales across the in-domain Test$_{in}$ (left column) and the out-of-domain Test$_{out}$ (center column) and Test$_{private}$ (left column) sets for the SB model. Different rows correspond to different anatomical scales indicated with icons on the left. The voxel-scale DSC-RCs and lesion-scale LPPV-RCs were obtained by averaging across the respective datasets. At each of the scales, the ideal (black dashed) line indicates the upper bound of an uncertainty measure performance in its ability to capture model errors; the random (gray dashed) indicates no relationship between an uncertainty measure and error; a worse-than-random performance indicates an inverse relationship.}
    \label{fig:erc-sb}
\end{figure}

Error retention curves for the SB model are shown in Figure \ref{fig:erc-sb}.

\subsection{Patient-scale uncertainty as a proxy for segmentation quality}
\label{appendices:corr-pat}

\begin{figure}[h!]
    \centering
    \begin{subfigure}[b]{0.77\linewidth}  
        \centering
        \caption{SB model}
        \includegraphics[width=\linewidth]{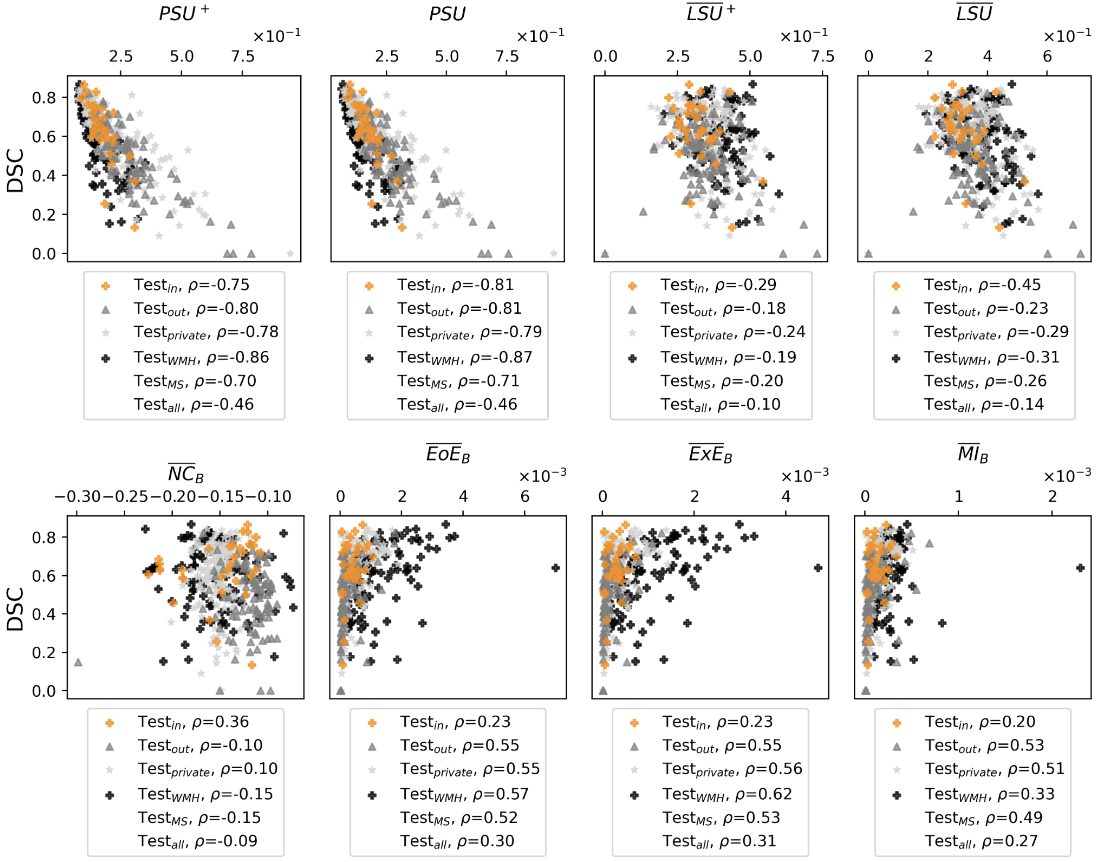} 
        
        \label{fig:7-sb}
    \end{subfigure}
    \hfill
    \begin{subfigure}[b]{0.77\linewidth}
        \centering
        \caption{nnU-Net model}
        \includegraphics[width=\linewidth]{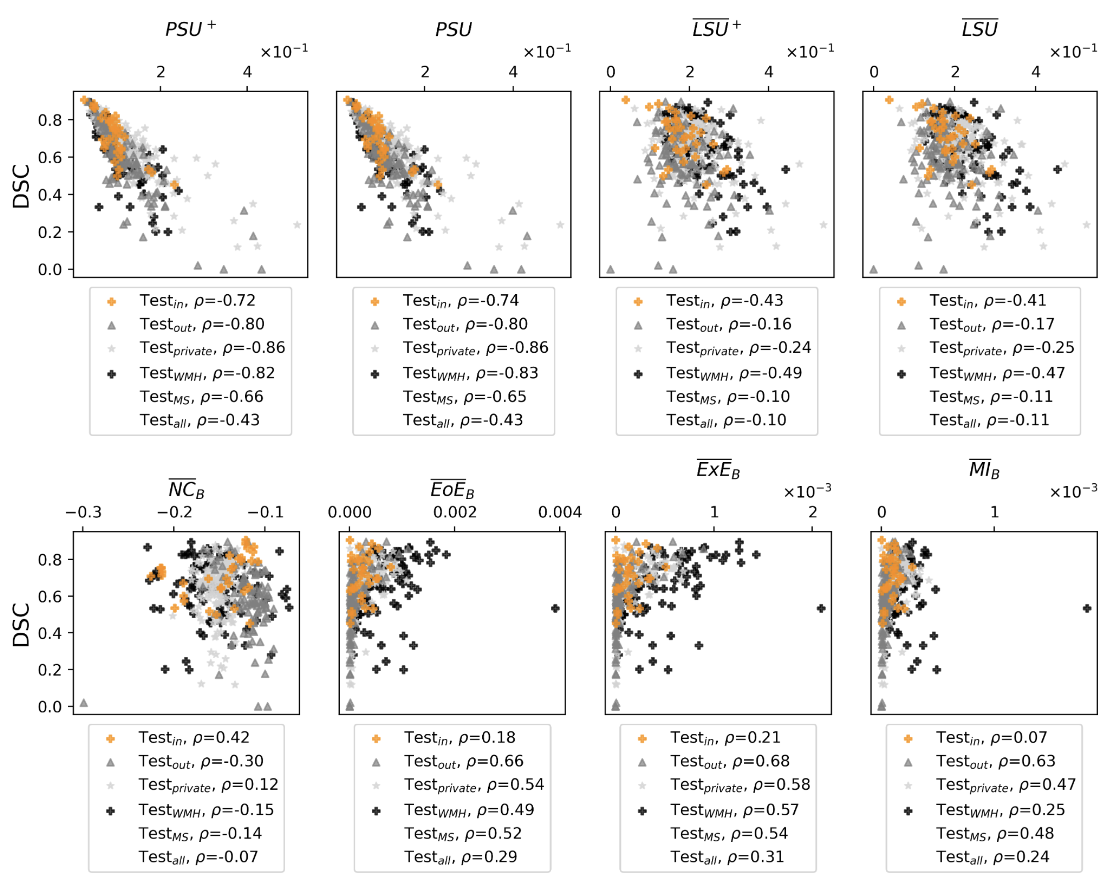}        
        \label{fig:8-nn}
    \end{subfigure}
    
    \caption{
    The relationship between the total ground truth lesion volume in milliliters (logarithmic y-axis) and various patient uncertainty measures (x-axis). \( \rho \) (in the legend) is a Spearman's correlation coefficient. 
    }
    \label{fig:78}
\end{figure}

Figure \ref{fig:78} extend the error retention curves analysis of the patient-scale uncertainty measures revealing more information about the relationship between the uncertainty measures and the segmentation quality measures by DSC.




\subsection{Generalizability analysis for white matter hyperintensity (WMH)}
\label{appendix:generalizability}

Areas under error retention curves for different anatomical scales are shown in Table \ref{tab:auc-wmh}.

\subsection{Uncertainty relationship with lesion size and load}
\label{appendix:load}

Lesion-scale analysis of the relationship between the predicted lesion volumes and uncertainty are shown in violin plots in Figures 9 a) and b) for SB and nnU-Net models, respectively. For all the lesion-scale uncertainty measures, lesions with smaller sizes tend to be more uncertain. For the nnU-Net model, the difference in medians of proposed $LSU^{(+)}$ uncertainty across different lesion volumes is less prominent compared to other measures.

 Patient-scale analysis of the relationship between the ground-truth total lesion volume and patient-scale uncertainty measure is given in Figure 10 a) and b) for SB and nnU-Net models, respectively. Different measures have a different degree of associations with the ground-truth total lesion volume (TLV):

 \begin{itemize}
     \item $PSU^{(+)}$ values are negatively associated with the TLV, meaning that a patient with low uncertainty is more likely to have a high TLV;
    \item Average $\overline{LSU^{(+)}}$ and $\overline{NC}_{B}$ show a mild negative association with the TLV;
    \item The rest of the aggregated voxel-scale measures ($\overline{EoE}_{B}$, $\overline{ExE}_{B}$ and $\overline{MI}_{B}$) have a strong positive association with the TLV: higher uncertainty for subjects with the higher TLV. This should explain a poor relationship with the overall segmentation quality, which tends to be higher for the patients with higher lesion loads.
 \end{itemize}

\begin{table}[h!]
\centering
\small
\begin{tabular}{|c|c|c|}
\hline
\textbf{Measure} & \textbf{SB} & \textbf{nnU-Net} \\ \hline
\multicolumn{3}{|c|}{\textbf{Voxel-scale DSC-AUC ($\uparrow$)}} \\ \hline
\textbf{Ideal}   & 99.81 [99.76, 99.83] & 99.85 [99.80, 99.88] \\
\textbf{$NC_i$}  & 98.41 [98.11, 98.62] & 99.40 [99.23, 99.50] \\
\textbf{$EoE_i$} & 98.35 [98.05, 98.59] & 99.40 [99.23, 99.51] \\
\textbf{$ExE_i$} & 98.35 [98.05, 98.59] & 99.39 [99.22, 99.50] \\
\textbf{$MI_i$}  & 98.16 [97.84, 98.41] & 99.10 [98.97, 99.46] \\
\textbf{Random}  & 76.11 [74.06, 77.85] & 80.15 [78.28, 81.70] \\ \hline
\multicolumn{3}{|c|}{\textbf{Lesion-scale LPPV-AUC ($\uparrow$)}} \\ \hline
\textbf{Ideal}   & 79.17 [75.94, 81.82] & 91.98 [89.95, 93.48] \\
\textbf{$LSU$}     & 73.32 [70.13, 76.12] & 86.64 [84.37, 88.43] \\
\textbf{$LSU^+$} & 73.03 [69.87, 75.83] & 86.64 [84.35, 88.44] \\
\textbf{$\overline{NC}_L$}  & 72.90 [69.76, 75.67] & 86.81 [84.51, 88.61] \\
\textbf{$\overline{EoE}_L$} & 72.94 [69.80, 75.71] & 86.80 [84.52, 88.59] \\
\textbf{$\overline{ExE}_L$} & 68.77 [65.72, 71.59] & 85.27 [82.89, 87.17] \\
\textbf{$\overline{MI}_L$}  & 72.98 [69.79, 75.70] & 86.70 [84.44, 88.50] \\ \hline
\textbf{Random}  & 66.38 [63.38, 69.08] & 82.00 [79.72, 83.91] \\ \hline
\multicolumn{3}{|c|}{\textbf{Patient-scale DSC-AUC ($\uparrow$)}} \\    \hline
\textbf{Ideal}   & 84.17 [82.99, 85.32] & 86.69 [85.62, 87.69] \\
\textbf{$PSU$}     & 83.51 [82.18, 84.75] & 85.92 [84.62, 87.05] \\
\textbf{$PSU^+$} & 83.47 [82.14, 84.72] & 85.86 [84.56, 86.98] \\
\textbf{$\overline{LSU}$}     & 81.09 [79.82, 82.29] & 84.60 [83.49, 85.70] \\
\textbf{$\overline{LSU}^+$} & 80.59 [79.29, 81.80] & 84.69 [83.59, 85.77] \\
\textbf{$\overline{NC}_B$}  & 80.06 [78.34, 81.53] & 82.88 [81.19, 84.25] \\
\textbf{$\overline{EoE}_B$} & 77.16 [75.83, 78.44] & 80.66 [79.41, 81.84] \\
\textbf{$\overline{ExE}_B$} & 76.93 [75.59, 78.22] & 80.28 [79.00, 81.47] \\
\textbf{$\overline{MI}_B$}  & 78.27 [76.90, 79.56] & 81.76 [80.49, 82.93] \\
\textbf{Random}  & 78.82 [77.10, 80.39] & 81.76 [80.06, 83.20] \\ \hline
\end{tabular}
\caption{
Mean average areas under error retention curves and 90\% bootstrap confidence intervals for the assessment of the uncertainty measures at the voxel, lesion, and patient anatomical scales across the WMH Challenge dataset (Test$_{WMH}$). Results are presented for the Shifts Challenge Baseline (SB) and nnU-Net models. Highest AUC values for each dataset, model, and anatomical scale are highlighted in bold, lowest - in italic; ideal and random values are in gray color and indicate the upper and lower bounds of performance, respectively.
}
\label{tab:auc-wmh}
\end{table}

\begin{figure}[h!]
    \centering
    \begin{subfigure}[b]{0.7\linewidth}  
        \centering
        \caption{SB model}
        \includegraphics[width=\linewidth]{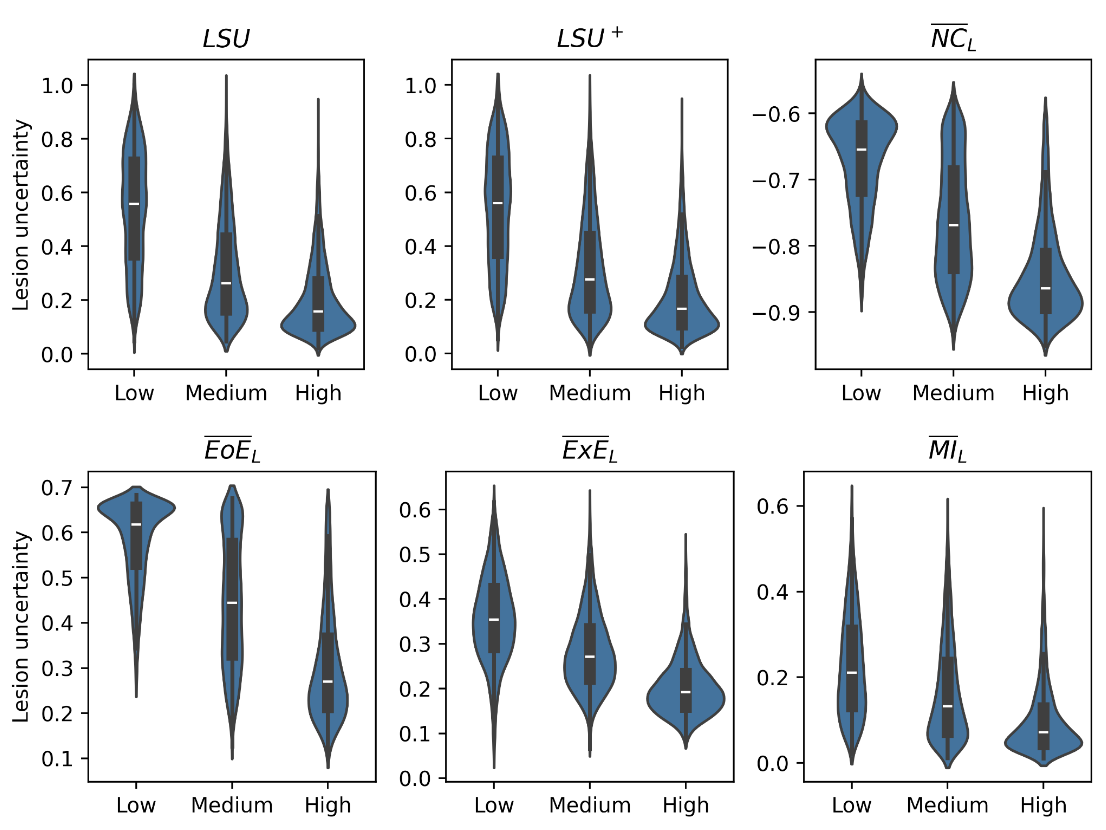} 
        
        \label{fig:vol-sb}
    \end{subfigure}
    \hfill
    \begin{subfigure}[b]{0.7\linewidth}  
        \centering
        \caption{nnU-Net model}
        \includegraphics[width=\linewidth]{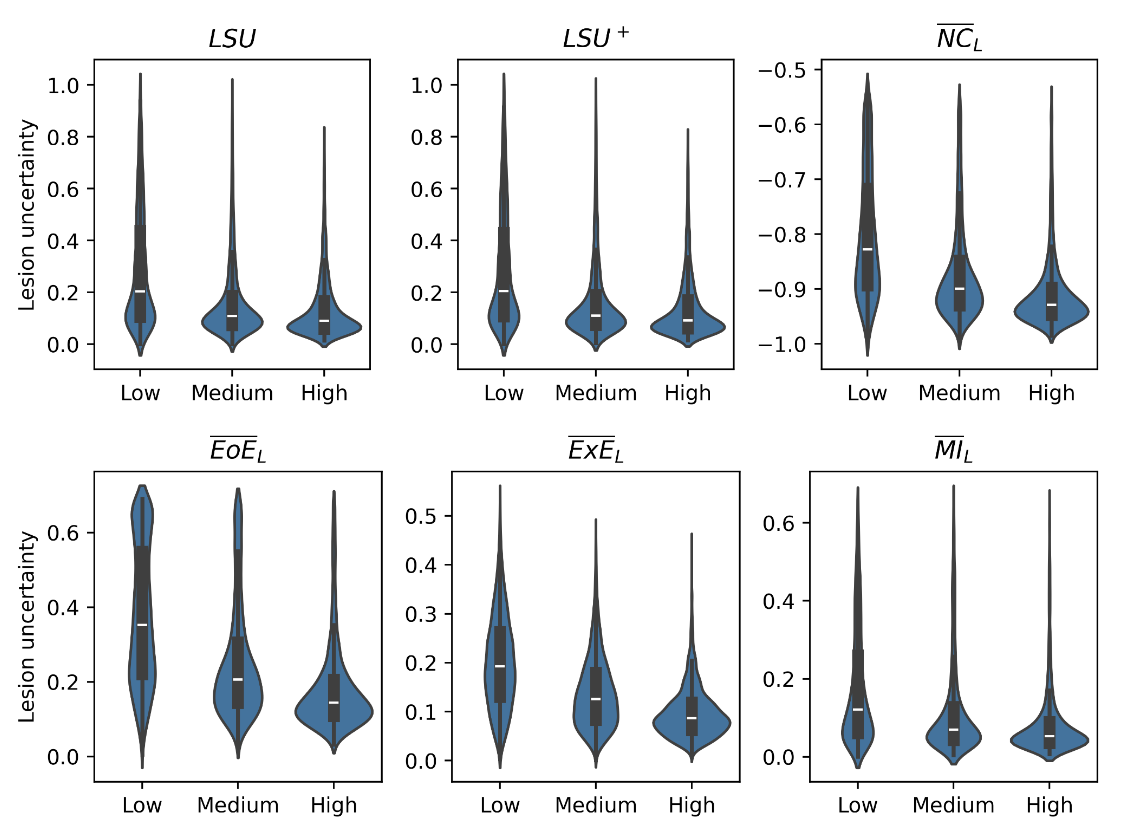} 
        
        \label{fig:vol-nn}
    \end{subfigure}
    
    \caption{
    The distributions of lesion uncertainty across 3 groups of predicted lesions in all the test sets jointly (Test$_{in}$, Test$_{out}$, Test$_{private}$, Test$_{WMH}$) defined through their volume percentiles: Low (0\%-33\%), Medium (33\%-67\%), High (67\%-100\%). 
    }
    \label{fig:vol}
\end{figure}

\begin{figure}[h!]
    \centering
    \begin{subfigure}[b]{0.77\linewidth}  
        \centering
        \caption{SB model}
        \includegraphics[width=\linewidth]{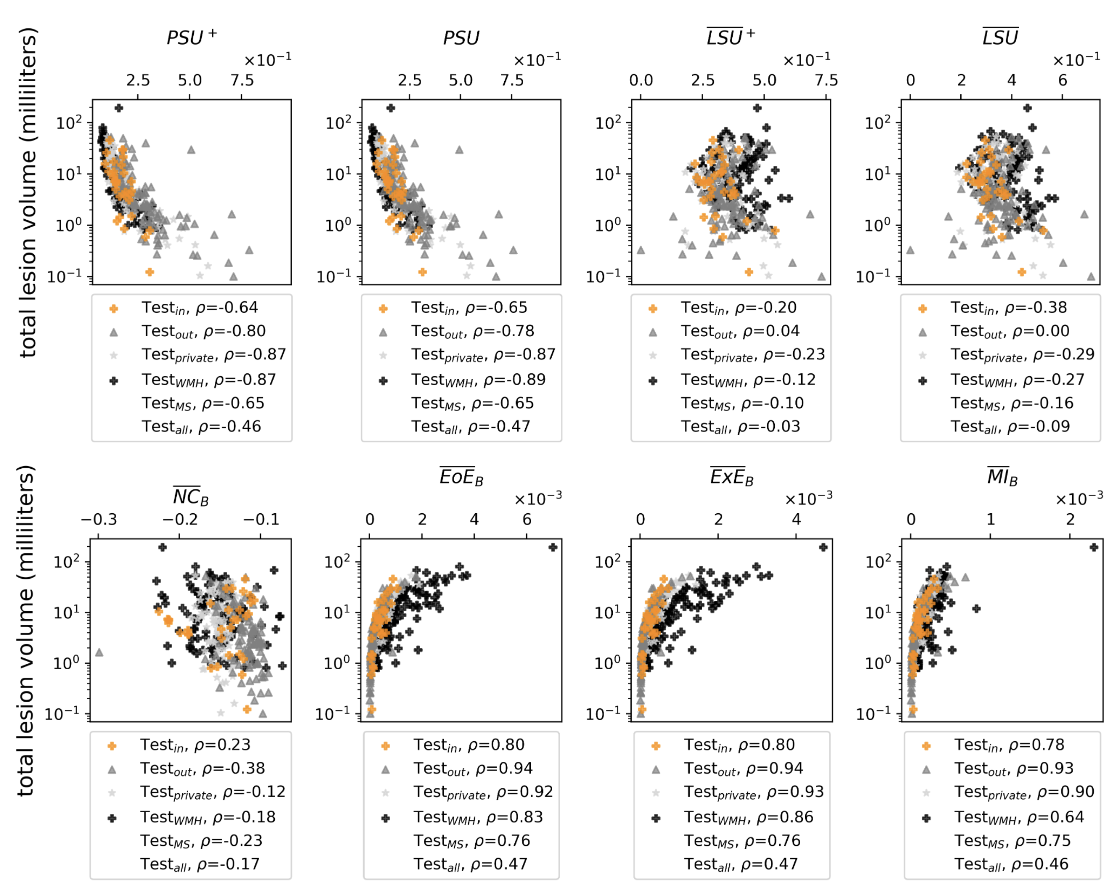} 
        
        \label{fig:tlv-sb}
    \end{subfigure}
    \hfill
    \begin{subfigure}[b]{0.77\linewidth}
        \centering
        \caption{nnU-Net model}
        \includegraphics[width=\linewidth]{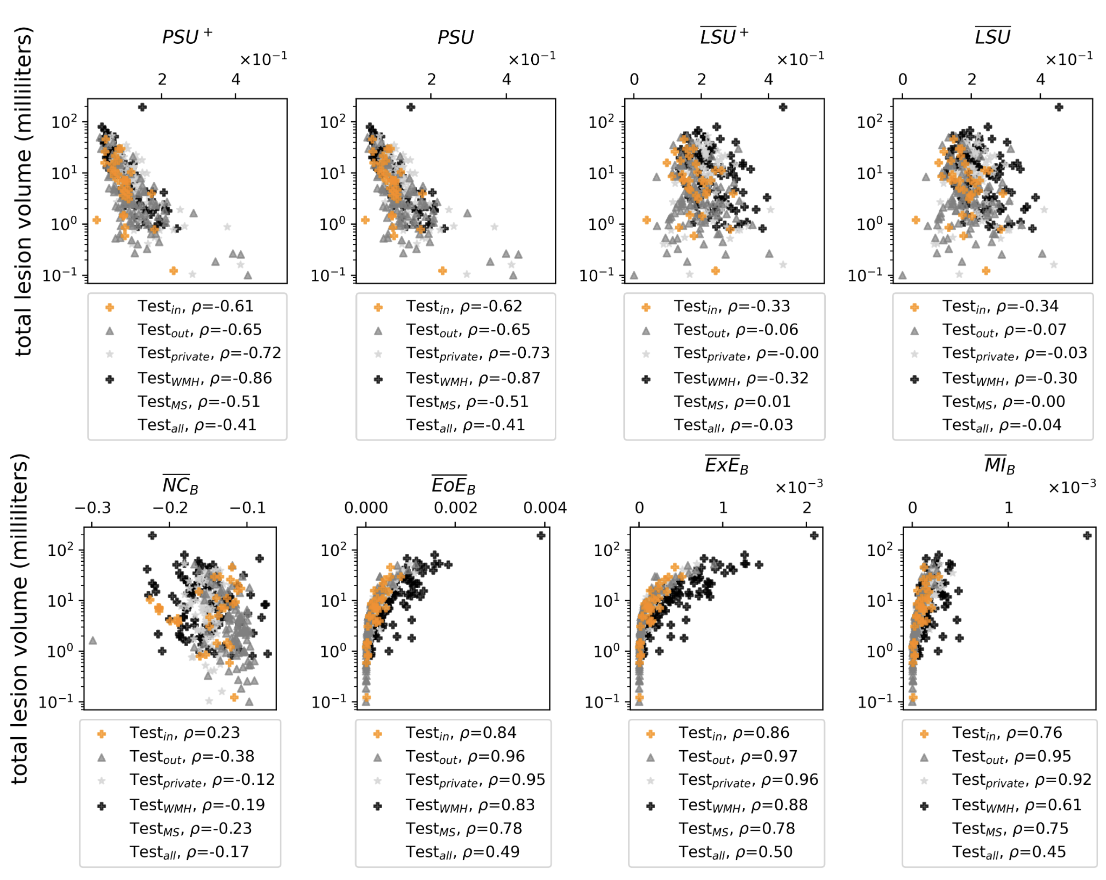}        
        \label{fig:tlv-nn}
    \end{subfigure}
    
    \caption{
    The relationship between the total ground truth lesion volume in milliliters (logarithmic y-axis) and various patient uncertainty measures (x-axis). \( \rho \) (in the legend) is a Spearman's correlation coefficient. 
    }
    \label{fig:tlv}
\end{figure}